\documentclass[journal]{IEEEtran}
\ifCLASSINFOpdf
\else
\fi
\newcommand\at[2]{\left.#1\right|_{#2}}
\usepackage{amsmath}
\usepackage{amssymb}

\usepackage{graphicx}
\usepackage{xcolor}
\usepackage{color}
\newcommand{\PK}{\color{black}}
\newcommand{\MH}{\color{black}}

\usepackage{subcaption}

\begin{document}
%
\title{Impact of Nonlinear Power Amplifier on Massive MIMO: Machine Learning Prediction Under Realistic Radio Channel}
%
%
%
\author{Marcin~Hoffmann,~\IEEEmembership{Member,~IEEE,}
        Pawel~Kryszkiewicz,~\IEEEmembership{Senior Member,~IEEE,}
\thanks{Copyright (c) 2026 IEEE. Personal use of this material is permitted. However, permission to use this material for any other purposes must be obtained from the IEEE by sending a request to pubs-permissions@ieee.org. 

This research was funded by the Polish National Science Centre, project no. 2021/41/B/ST7/00136, and Polish Ministry of Science and Higher Education, project no. 0312/SBAD/8170.
Authors are with Institute of Radiocommunications, Poznan University of Technology, Poznan, Poland, and with Rimedo Labs, Poznań Poland. 
Corresponding author: M. Hoffmann, e-mail: marcin.hoffmann@put.poznan.pl}}

%
%

\markboth{Journal of \LaTeX\ Class Files,~Vol.~14, No.~8, August~2015}%
{Shell \MakeLowercase{\textit{et al.}}: Bare Demo of IEEEtran.cls for IEEE Journals}
%



\maketitle

\begin{abstract}

Massive Multiple-Input Multiple-Output (M-MIMO) is one of the crucial technologies for increasing spectral and energy efficiency of wireless networks. Most of the current works assume that M-MIMO arrays are equipped with a linear front end. However, ongoing efforts to make wireless networks more energy-efficient push the hardware to the limits, where its nonlinear behavior appears. This is especially a common problem for the multicarrier systems, e.g., Orthogonal Frequency Division Multiplexing (OFDM) used in 4G, 5G, and possibly also in 6G, which is characterized by a high Peak-to-Average Power Ratio (PAPR). While the impact of a nonlinear Power Amplifier (PA) on an OFDM signal is well characterized, it is a relatively new topic for the M-MIMO OFDM systems. Most of the recent works either neglect nonlinear effects or utilize simplified models proper for Rayleigh or Line of Sight (LoS) radio channel models. In this paper, we first theoretically characterize the nonlinear distortion in the M-MIMO system under commonly used radio channel models. Then, utilizing 3D-Ray Tracing (3D-RT) software, we demonstrate that these models are not very accurate. Instead, we propose two models: a statistical one and a Machine Learning (ML)-based one using 3D-RT results. The proposed statistical model utilizes the Generalized Extreme Value (GEV) distribution to model Signal to Distortion Ratio (SDR) for victim users, receiving nonlinear distortion, e.g., as interference from neighboring cells. The proposed ML model aims to predict SDR for a scheduled user (receiving nonlinear distortion along with the desired signal), based on the spatial characteristics of the radio channel and the operation point of each PA feeding at the M-MIMO antenna array. The predicted SDR can then be used to perform PA-aware per-user power allocation. The results show about 12\% median gain in user throughput achieved by the proposed ML-based power allocation scheme over the state-of-the-art, fixed operating point scheme.
\end{abstract}


%
\IEEEpeerreviewmaketitle

\section{Introduction}
\label{sec_introduction}
%
%
%
%

\IEEEPARstart{T}{he} Massive Multiple-Input Multiple-Output (M-MIMO) is one of the crucial technologies for both currently deployed 5G and future 6G networks~\cite{Yang2019}. It utilizes large antenna arrays to transmit signals directly toward User Equipments (UEs), improving spectral efficiency and enabling spatial division multiple access. Most of the current works assume that M-MIMO arrays are equipped with a linear front end. This is somehow a simplification enforced by the wireless standards, i.e., the Power Amplifiers (PAs) deployed at the Base Stations (BSs) usually operate in their linear region to avoid signal clipping that is typically enforced by constraining an Error Vector Magnitude (EVM) at the transmitter \cite{3gpp38104}. However, current efforts to make 6G networks more energy-efficient push the hardware to the limits, where its nonlinear behavior appears~\cite{Perre2018}. This is a common problem for the multicarrier systems, e.g., Orthogonal Frequency Division Multiplexing (OFDM) used in 4G, 5G, and possibly also in 6G, which is characterized by a high Peak-to-Average Power Ratio (PAPR). While in general the distortion-less system seems like a better solution, in some cases more spectrally efficient to increase the distortion power if the wanted signal power increases as well {\PK as has been shown for the SISO system in \cite{tavares2016} and for the M-MIMO system under iid Rayleigh channel in \cite{Marwaha_TWC_2025}}. This idea can also be extended to energy efficiency maximization as investigated recently by 3GPP \cite{3GPP38864}. 

While the impact of a nonlinear PA on an OFDM signal is well characterized~\cite{gharaibeh2011nonlinear,lee2014characterization,Guerreiro_less_IMD_2015}, it is a relatively new topic for the M-MIMO systems. The main issue is how the nonlinear distortion signal generated at each transmitter chain adds in each spatial location, i.e., if the signals are uncorrelated and have joint omnidirectional characteristics, or if the distortion is correlated, creating some maxima at specific spatial locations.
The initial studies, like \cite{bjornson2014massive}, assumed an uncorrelated distortion, resulting in the wanted signal power increasing with the number of transmitting antennas, as a result of array gain, while the distortion signal simply adds as some kind of independent white noise generated by each antenna. However, in work~\cite{Larsson_2018_nonlinear_Los} it has been shown that the nonlinear distortion can achieve a similar array gain as the wanted signal, e.g., in the Line of Sight (LoS) channel, making it a significant issue that needs detailed analysis.  
Recently, authors in \cite{Teodoro_2019_analog_MIMO_PSD} modeled the influence of nonlinear PA on a hybrid M-MIMO OFDM system and its reception performance. While per front-end nonlinearity distortion is modeled accurately, the model does not consider the potential correlation of distortions from multiple transmit antennas at the reception point. As such, the nonlinear distortion can be underestimated in many cases. In \cite{mollen_spatial_char_2018}, the authors proposed a generalized framework to calculate the wanted signal and nonlinear distortion Power Spectral Density (PSD) in an M-MIMO system. While it considers the potential correlation between nonlinearity distortion signals between front-ends, the method is modulation agnostic (works both for multi-carrier and single-carrier signals) and considers a general memory polynomial for PA modeling. Most importantly, this is a semi-analytic method that evaluates PSDs for a single-channel realization, without showing a general relation between channel properties and distortion signal directivity. A recent study in \cite{Bjorson_2023_nonlinearities} utilized an analytical model from \cite{mollen_spatial_char_2018} and derived Signal to Distortion Ratio (SDR) for a M-MIMO transmission to a single user utilizing Maximal Ratio Transmission (MRT) precoding in OFDM signal. A simplification of PA modeling by only 3rd order polynomial was used. In addition, it was assumed that the channels are independent between antennas, but described by an impulse response of limited duration, i.e., there can be a correlation between channels on adjacent subcarriers for a single antenna if the impulse response is short enough. It has been shown that as the impulse response prolongs the distortion becomes more omnidirectional. While the result is interesting, the authors assumed a uniform distribution of power of paths over the channel profile, which is not typical \cite{3gpp38104}. Moreover, the authors did not consider varying mean transmit power over antenna elements, which can be a significant problem, as will be shown here.

In this paper, we analyze the impact of nonlinear PA on the M-MIMO OFDM transmission under a realistic channel. First, we provide a theoretical analysis of the nonlinear distortion under the commonly considered radio channel models: uncorrelated Rayleigh and LoS radio channels. We start the theoretical analysis with general formulas for the multiuser MIMO, and then, we focus on the case when a M-MIMO cell serves one UE, i.e., Multiple-Input Single-Output (MISO). The MISO is well established in the literature, e.g., works~\cite{Moothedath2020, Zhang2010}. Moreover, from the perspective of the nonlinear distortion, MISO is the most challenging scenario, where potentially the nonlinear distortion can be steered toward the UE~\cite{mollen_spatial_char_2018}. 
After providing the analytical formulas, we compare the results in these two specific channels against the results obtained with the use of a Wireless InSite 3D Ray-Tracing (3D-RT) software, considering a MISO system in an urban environment. The analysis is split between \emph{victim UEs} (receiving nonlinear distortion, e.g., as interference from neighboring cells) and \emph{scheduled UEs} (receiving nonlinear distortion along with the desired signal). 
In both cases, we show that nonlinear distortion power under the realistic radio channels affected by the spatial correlations is neither uncorrelated Rayleigh nor LoS, and follows a nontrivial distribution, i.e., the performed distribution fitting resulted in a closest match with the Generalized Extreme Value (GEV) distribution. For \emph{victim UE}, we propose a statistical model that models the SDR for a given Input Back Off (IBO), a measure of PA operating point, by the GEV distribution multiplied by the theoretical value of SDR. 
Moreover, while spatially adjacent locations exhibit similar SDR values, a decorrelation distance has been modeled. For a \emph{scheduled UE}, apart from the SDR showing high spatial variability and non-trivial distribution, we show that LoS is not the worst case for nonlinear distortion. The SDR can go below the LoS theoretical results when the power is allocated between antennas in a highly non-uniform manner.  We believe a correct estimation of nonlinear distortion power at each location is crucial for spectrally and energy-efficient resource allocation. We propose to utilize a Machine Learning (ML) method, namely a Convolution Neural Network (CNN), to predict SDR based on a modified channel correlation matrix. The modification is needed to embed information about the IBO associated with each PA. Finally, we propose to utilize the SDR prediction for the distortion-aware per-user power allocation, i.e., a method that will predict for a given UE a value of IBO to increase its Signal to Noise and Distortion Ratio (SNDR) based on its radio channel coefficients. Through the computer simulations we show that the proposed distortion-aware per-user power allocation can significantly improve the UE rates compared to the commonly used fixed IBO scheme.

{\MH The key contribution of the paper can be summarized as follows:
\begin{itemize}
    \item We provide a theoretical analysis of M-MIMO systems with nonlinear distortion under uncorrelated Rayleigh and LoS models of the radio channel. First, we derive general formulas for multiuser MIMO and then go to the detailed analysis for the MISO case. Observed significant SDR differences between the two models, highlighting the need for channel-specific SDR estimation.
    \item Unlike state-of-the-art works (e.g.,~\cite{bjornson2014massive,Bjorson_2023_nonlinearities}), we assess the impact of nonlinear distortion in M-MIMO using a realistic system-level simulator based on the 3D-RT channel model for urban environments.
    \item For non-scheduled (\emph{victim UEs}), we show that the commonly used uncorrelated Rayleigh model is inaccurate for SDR modeling. In contrast to computationally extensive, and radio channel realization-dependent methods like~\cite{mollen_spatial_char_2018}, we propose a statistical SDR model based on the GEV distribution and spatial correlation distance, validated for both soft-limiter and Rapp PA models. 
    \item For \emph{scheduled UEs}, contrary to~\cite{Teodoro_2019_analog_MIMO_PSD}, we demonstrate that SDR is affected by the antenna channel correlation modeled by 3D-RT. We further show that unequal path losses across antennas lead to different IBO values, which may reduce SDR below the theoretical LoS bound claimed as the worst case in~\cite{mollen_spatial_char_2018}.
    \item For \emph{scheduled UEs}, we propose an SDR prediction method based on an adapted VGG16 architecture~\cite{simonyan2014very}. The model is validated on datasets generated with 3D-RT and shown to deal with different PA characteristics (soft-limiter and Rapp). Moreover, hardware complexity is analyzed, and pruning is applied to reduce model size and inference time.
    \item We propose a distortion-aware per-user power allocation for M-MIMO that uses VGG16-based SDR prediction to dynamically adjust the IBO per UE. Simulation results based on 3D-RT demonstrate superior data rate performance compared to fixed-IBO schemes and dynamic IBO selection proposed in~\cite{tavares2016}.
\end{itemize}
}

The rest of the paper is organized as follows. Sec.~\ref{sec:system_model} provides a description of the system model with a focus on the theoretical foundations of nonlinear distortions in a M-MIMO OFDM system. Sec.~\ref{sec:ray-tracer-analysis} presents the 3D-RT-based analysis of the nonlinear distortion for both \emph{victim UEs} and \emph{scheduled UEs}, together with comparison to the theoretical results, and important observations. The ML-based SDR prediction, together with the proposed distortion-aware per-user power allocation, is provided in Sec.~\ref{sec:ml_based_sdr_predioction}. Finally, the conclusions are formulated in Sec.~\ref{sec:conclusions}




\section{System Model} \label{sec:system_model}
\begin{figure}[!t]
\centering
\includegraphics[trim={0cm 2cm 0cm 0cm}, clip,width=0.49\textwidth]{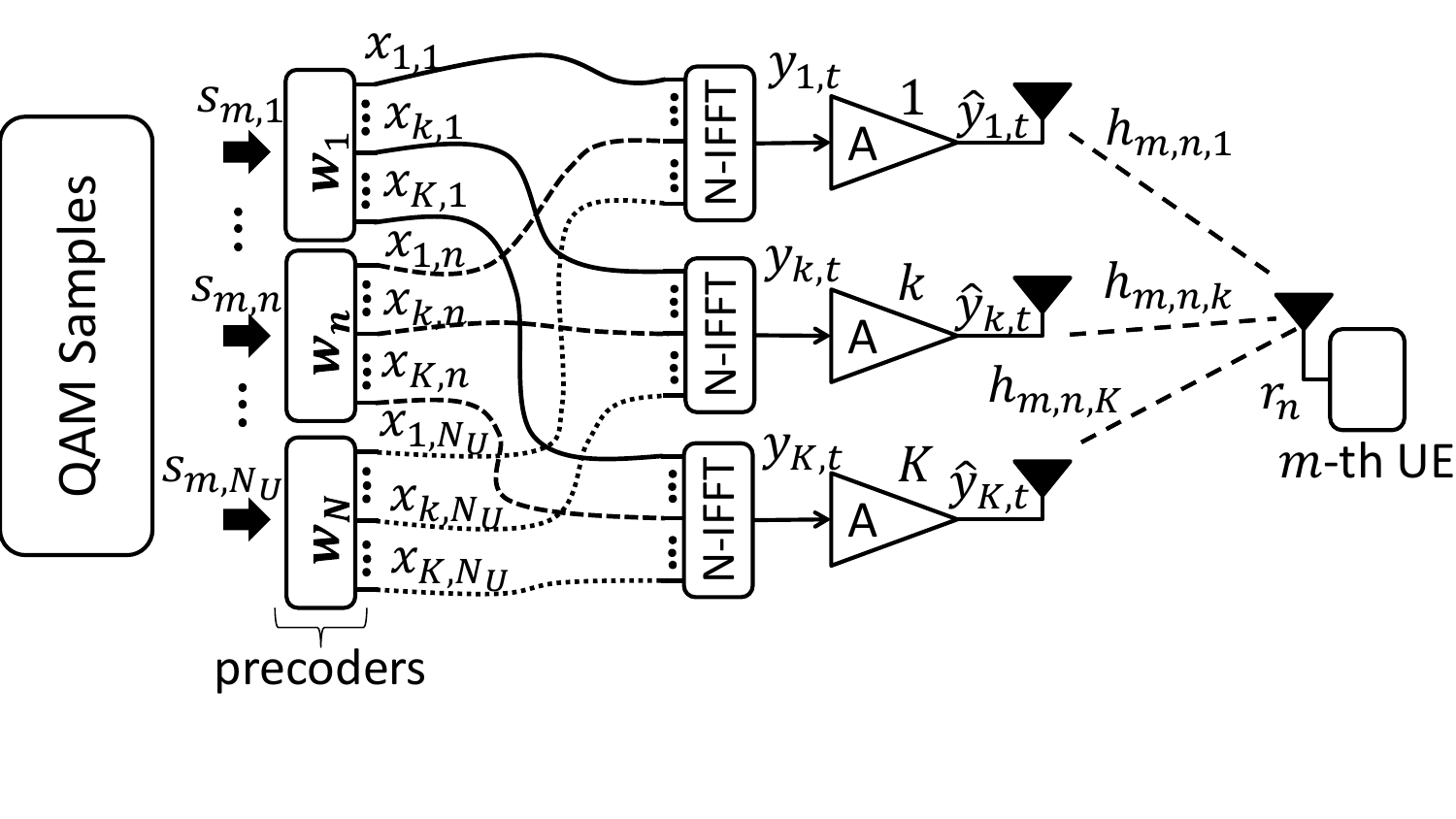}
\caption{System model.}
\vspace{0.25cm}
\label{fig:system_model}
\end{figure}

We are considering a downlink in a single MISO OFDM base station (BS) operating at carrier frequency $f_c$. The related system model is depicted in Fig.~\ref{fig:system_model}.
The BS is equipped with $K$ transmitter chains, each including a digital precoder, OFDM modulator, and radio front end. There are $M$ User Equipments (UEs), each using a single antenna, served at a given time. 
The OFDM modulator utilizes $N$-point Inverse Fast Fourier Transform (IFFT) block with $N_{u}\leq N$ subcarriers modulated with non-zero Quadrature Amplitude Modulation (QAM)/ Phase Shift Keying (PSK) symbols. The symbol intended for $m$-th UE ($m\in \{1,...,M \}$) on $n$-th subcarrier ($n\in \{1,...,N_{\mathrm{U}} \}$) is denoted as $s_{m,n}$ that will be transmitted during a period of a single OFDM symbol. The $n$-th QAM/PSK symbol is transmitted on subcarrier $I_n \in \{-N/2,...,N/2-1 \}$ of the OFDM symbol. 
We assume that the radio channel between the $k$-th transmitter antenna and $m$-th UE is flat within the bandwidth of a single subcarrier and does not change over the period of a single OFDM symbol. This channel's impulse response is denoted by $h_{m,n,k}$.
 First, the complex data symbols undergo digital precoding, creating symbol $x_{k,n}$ as
\begin{equation}
\label{eq:precoding}
    x_{k,n} = \sum_{m}s_{m,n} w_{m,n,k},
\end{equation}
where $w_{m,n,k}$ is a precoding coefficient. One of the possible approaches to designing the precoder is the Maximum Ratio transmission (MRT) scheme, which maximizes the received signal power~\cite{bjornson2017}:
\begin{equation}
\label{eq:mrt_precoder}
w_{m,n,k} = \frac{h^*_{m,n,k}}{\sqrt{\sum_{k}\sum_{m} \left| h_{m,n,k} \right|^2}},    
\end{equation}
where $^*$ denotes complex conjugate, and the normalization introduced in the denominator of the above equation ensures $\sum_k|{w_{m,n,k}}|^2=1$. 
It is assumed that the same normalization is used for any precoding scheme applied.

After the precoding stage, the resultant $x_{k,n}$ samples are subject to the OFDM modulation~\cite{haykin2013digital} obtaining $t$-th sample of time domain signal intended for the $k$-th antenna:
\begin{equation}\label{eq:ofdm}
    y_{k,t} = \sum_{n} x_{k,n}e^{j2\pi \frac{I_n}{N}t}, 
\end{equation}
where $t \in \{-N_\mathrm{CP}, \dots, N - 1\}$, and $N_\mathrm{CP}$ is the number of samples of the cyclic prefix (CP). 
Next, the signal is processed through the digital-to-analog converter and upconverted to the chosen carrier frequency. 
The final stage of signal processing at the transmitter side is a PA. We consider the same PA to be used by each of the $K$ transmitter chains. The behavioural PA model, i.e., from a signal path perspective\cite{gharaibeh2011nonlinear}, is described by a nonlinear function $\mathcal{A}$. As a result, the equivalent (modeled at baseband) signal transmitted by the $k$-th antenna is given by: 
\begin{equation}
    \hat{y}_{k,t} = \mathcal{A}(y_{k,t}).
\end{equation}
There are multiple PA models in the literature~\cite{gharaibeh2011nonlinear}. {\PK In \cite{ochiai2013analysis} a Rapp model has been suggested to characterize a typical solid state PA used in modern cellular systems, i.e., 
\begin{equation}\label{eq:rapp_pa}
    \hat{y}_{k,t} =\frac{y_{k,t}}{\left(1+\frac{|y_{k,t}|^{2p}}{P_{\mathrm{max},k}^{p}}\right)^{\frac{1}{2p}}},
\end{equation}
where $p$ is the smothness parameter of a PA and $P_{\mathrm{max},k}$ is the maximum possible output sample power of the $k$-th frontend PA, i.e., the saturation power. While $P_{\mathrm{max}, k}$ is assumed here to be fixed in time, there are approaches like Envelope Tracking that can vary $P_{\mathrm{max}, k}$\cite{Asbeck2016}. While \cite{ochiai2013analysis} suggests $p=2$ as a typical value, when $p\to \infty$ the so-called soft-limiter PA is obtained. The soft-limiter is an ultimate PA design goal, as it is proven to be a nonlinearity that maximizes the SDR of an OFDM waveform~\cite{raich2005}. The soft-limiter clips the input signal if its power exceeds the maximum transmit power of the $k$-th front-end PA, otherwise the input signal passes through the PA unchanged, i.e.,
\begin{equation}\label{eq:soft_limiter}
    \hat{y}_{k,t} = \begin{cases} y_{k,t} &\text{if} \hspace{0.1cm} |y_{k,t}|^2 \leq P_{\mathrm{max}, k} \\
    \sqrt{P_{\mathrm{max},k}}e^{j \arg(y_{k,t})} &\text{otherwise}
    \end{cases},
\end{equation}
where the $\arg(y_{k,t})$ denotes the phase of $y_{k,t}$. 

Even if the PA is characterized by a more severe nonlinear characteristics, including those revealing memory effects \cite{gharaibeh2011nonlinear}, typically the modern transmitters are equipped with a digital predistortion (DPD) block at the PA input \cite{Joung_PA_centric_survey_2015}.  This should result in the combined DPD and the PA characteristics close to the soft-limiter characteristics. 
}
Moreover, while \cite{mollen_spatial_char_2018} analyzed the influence of memory polynomial PA on the M-MIMO system, it requires knowledge of multiple polynomial coefficients that typically change with the operating point of the amplifier \cite{lee2014characterization}.
{\PK In most parts of this paper}, the soft-limiter assumption will be used, allowing for easier results reuse in further studies. However, the nonlinear distortion evaluation and ML modeling framework shown in the next sections can be easily adapted to any memoryless PA characteristics reported by a wireless transmitter vendor.



Observe that the nonlinear distortion will be, in general, the more severe the more samples are clipped. While the precoded OFDM symbol samples should be characterized by a zero-mean, complex-Gaussian distribution, as a result of the addition of a large number of independent subcarriers~\cite{Wei_2010_dist_OFDM}, the statistical characterization of the transmitted signal distribution depends only on the mean signal power. 
Let us define the transmit power at the PA input per $k$-th antenna, by using (\ref{eq:precoding}) and (\ref{eq:ofdm}) as
\begin{equation}
\label{eq_pk_def}
    p_k=\sum_n\sum_m |w_{m,n,k}|^2 \mathbb{E}[|s_{m,n}|^2],
\end{equation}
where $\mathbb{E}[~]$ denotes expectation. In the above, it is assumed that the symbols between subcarriers and UEs are uncorrelated. 
Usually, its relation with the PA clipping level is defined by an Input Back-Off (IBO) equal
\begin{equation} \label{eq:ibo_k}
\gamma_k = \frac{P_\mathrm{max,k}}{p_k}.
\end{equation}
While the above, per-antenna metric is more meaningful, especially for a precoding unequally distributing power among antennas, an average IBO can be defined as:
\begin{equation} \label{eq:ibo}
\gamma = \frac{\sum_k P_{\mathrm{max},k}}{\sum_k p_k}.
\end{equation}

While the complex-Gaussian OFDM signal passes through a {\PK memoryless} nonlinear system, a Bussgang theory can be applied, allowing us to decompose the distorted signal into \cite{Nossek2011power,kryszkiewicz2023efficiency}
\begin{equation}
\label{eq_bussgang}
    \hat{y}_{k,t}=\lambda_k y_{k,t} +\hat{d}_{k,t},
\end{equation}
where $\hat{d}_{k,t}$ is the distortion signal, uncorrelated with the wanted signal, and $\lambda_k\in (0,1)$ is a scaling factor that is defined as
\begin{equation}
    \lambda_k\!=\!\frac{
\mathbb{E} \left[ \at{\hat{y}_{\hat{k},t}y_{\hat{k},t}^{*}}{\hat{k}=k} \right]
    }{\mathbb{E} \left[ \left| \at{y_{\hat{k},t}}{\hat{k}=k} \right|^2\right]}\!=\!1\!-\!e^{-\gamma_k}\!+\!\frac{1}{2}\sqrt{\pi \gamma_k} \mathrm{erfc}\left(\sqrt{\gamma_k}\right)\!,
    \label{eq_lambda}
\end{equation}
{\PK with the last part of the above equality valid for soft-limiter PA. However, the general definition of $\lambda_k$ can be used for any memoryless nonlinear PA.}
This allows us to define the wanted signal power over the $k$-th antenna as
\begin{equation}
    S_k^{\mathrm{TX}}=\lambda_k^{2}p_k
\end{equation}
observing that it accounts for all users' signals, i.e., inter-user (linear) interference as well. 
The distortion power can be calculated as \cite{kryszkiewicz2023efficiency}
 \begin{equation}
   D_k^{\mathrm{TX}}=\mathbb{E} \left[ |\at{\hat{d}_{\hat{k},t}}{\hat{k}=k}|^{2} \right]= \left(1-e^{-\gamma_k}-\lambda_k^2 \right)p_{k}. 
   \label{eq_per_UE_distortion_effective}
\end{equation}

The signal $\hat{y}_{k,t}$ is transmitted over the channel of response $\tilde{h}_{n,k}$ at subcarrier $n$ to a single antenna receiver. Observe that this channel can be equal to $h_{m,n,k}$ for some user $m$ or different, for a not-scheduled user. In the receiver, the synchronized signal undergoes $N$-point FFT with the output sample on $n$-th subcarrier equal
\begin{equation}
\label{rx_sig_freq_domain}
    r_{n}=\sum_{k} \mathcal{F}_{[n,t=0,...,N-1]}\{\hat{y}_{k,t}\}\tilde{h}_{n,k} + w_{n},
\end{equation}
where $\mathcal{F}_{[n,t=0,...,N-1]}$ denotes discrete Fourier transform (DFT) over time samples $0,...,N-1$ evaluated at subcarrier $n$, and $w_{n}$ is Additive White Gaussian Noise (AWGN) sample. After substitution of (\ref{eq:precoding}) and (\ref{eq_bussgang}) this sample can be decomposed as
\begin{subequations}
\begin{align}
    r_{n}=
   &\sum_m s_{m,n} \sum_k \lambda_k \tilde{h}_{n,k}w_{m,n,k}
    \label{rx_sig_freq_domain_decomp1}
    \\
    &+
    \sum_{k} d_{n,k}\tilde{h}_{n,k}\label{rx_sig_freq_domain_decomp3}
    \\
    &+ w_{n},\label{rx_sig_freq_domain_decomp4}
\end{align}
\end{subequations}
where
\begin{equation}
    d_{n,k}=\mathcal{F}_{[n,t=0,...,N-1]}\{\hat{d}_{k,t}\}.
\end{equation}
Observe that (\ref{rx_sig_freq_domain_decomp1}) can contain both wanted signal, if channel $\tilde{h}_{n,k}$ is to one of the $M$ scheduled UEs, and inter-user, linear interference. The component (\ref{rx_sig_freq_domain_decomp3}) denotes nonlinear distortion sample.
{\PK In general, for $m$-th user the wanted signal power at $n$-th subcarrier can be defined as
\begin{equation}
    S_n^{\mathrm{RX}}=\mathbb{E}\left[ 
    \left|s_{m,n} \sum_k \lambda_k \tilde{h}_{n,k}w_{m,n,k} \right|^2
    \right]
\end{equation}
and distortion power as 
\begin{equation}
    D_n^{\mathrm{RX}}=\mathbb{E}\left[ 
    \left|
   \sum_{k} d_{n,k}\tilde{h}_{n,k}
    \right|^2
    \right].
\end{equation}
This can be calculated by means of simulations for any memoryless PA by approximating the expectation by averaging over multiple transmitted OFDM symbols. Finally, the mean SDR can be calculated as
\begin{equation}
    SDR=\frac{\sum_{n} S_n^{\mathrm{RX}}}{\sum_n D_n^{\mathrm{RX}}}.
   \label{eq_SDR}
\end{equation}
}

In the case of a single UE scheduling ($M=1$) using MRT precoding, as in \cite{Bjorson_2023_nonlinearities}, and observing the signal at this UE, i.e., $\tilde{h}_{n,k}=h_{1,n,k}$, with the assumption of equal large scale fading per antenna, i.e., $\forall_{n,n',k,k'}\mathbb{E}[|\tilde{h}_{n,k}|^2]=\mathbb{E}[|\tilde{h}_{n',k'}|^2]$ and equal $P_{max,k}$ per antenna, the received signal power over all utilized subcarriers equals
\begin{equation}
    S^{\mathrm{RX}}=K \mathbb{E}[|\tilde{h}_{n,k}|^2] |\lambda|^2 \sum_k p_k
    \label{eq_wanted_RX_power}
\end{equation}
where $\mathbb{E}[|\tilde{h}_{n,k}|^2]$ denotes mean channel gain and $\lambda$ is used for $\lambda_k$ as a result of IBO equal over all frontends. This was obtained by calculating $\mathbb{E}[|\sum_m s_{m,n} \sum_k \lambda_k \tilde{h}_{n,k}w_{m,n,k}|^2]$ while utilizing (\ref{eq:mrt_precoder}) and  (\ref{eq_pk_def}). It is visible that, as in a standard MRT case, the total transmit power $\sum_k p_k$ obtains array gain $K$. The only difference is the reduction of the wanted signal power by a factor $|\lambda|^2$ as a result of nonlinear distortion.

The total distortion power over all utilized ($N_{\mathrm{U}}<N$) subcarriers, assuming the distortion is uncorrelated with the wireless channel, 
as well as the channel's large-scale fading equal over antennas (as above), can be calculated based on (\ref{rx_sig_freq_domain_decomp3}) as 
\begin{equation}
   D^{\mathrm{RX}}\!= \!\!\sum_n \!\mathbb{E} \!\left[ \left| \sum_{k} d_{n,k}\tilde{h}_{n,k}\right|^2 \right]
   \!\!=\!\!\mathbb{E}[|\tilde{h}_{n,k}|^2] \sum_k \sum_n \mathbb{E}[|d_{n,k}|^2].
\end{equation}
Observe that $\sum_n \mathbb{E}[|d_{n,k}|^2]$ is a total distortion power generated from a single front-end over all $N_u$ utilized subcarriers. Based on Parsevall theorem this should be equal to $D_k^{\mathrm{TX}}$ from (\ref{eq_per_UE_distortion_effective}), but some of the time-domain distortion signal will leak into the out-of-band (OOB) frequency region. In \cite{lee2014characterization}, it was derived that approximately $2/3$ of the total distortion signal is emitted within the utilized subcarriers range. {\PK While the 2/3 constant is accurate only for 3rd order intermodulation products, with the higher order intermodulations leaking even more significantly into the OOB region, in the usable IBO range, the 3rd order intermodulations dominate \cite{lee2014characterization}. Therefore the constant $2/3$ results in an accurate approximation as shown by simulations in \cite{Marwaha_TWC_2025}, significantly improving the time domain model used, e.g., in \cite{kryszkiewicz2023efficiency}.} This allows to write $\sum_n \mathbb{E}[|d_{n,k}|^2]=\frac{2}{3} D_k^{\mathrm{TX}}$.  
Finally, by utilizing the above equation and definition in (\ref{eq_per_UE_distortion_effective}), recalling that IBO at each front-end is identical and equals $\gamma$, we obtain
\begin{equation}
   D^{\mathrm{RX}}= \mathbb{E}[|\tilde{h}_{n,k}|^2] \frac{2}{3}  \left(1-e^{-\gamma}-\lambda^2 \right) \sum_k p_{k}. 
   \label{eq_distortion_uncorrelated}
\end{equation}
Using (\ref{eq_wanted_RX_power}) and (\ref{eq_distortion_uncorrelated}) the Signal to Distortion Ratio (SDR) for uncorrelated distortion case, e.g., IID Rayleigh channel \cite{bjornson2014massive}, can be calculated as
\begin{equation}
    SDR^{\mathrm{uncorrelated}}=\frac{K \lambda^2}{\frac{2}{3}\left(1-e^{-\gamma}-\lambda^2 \right)} .
   \label{eq_SDR_uncorrelated}
\end{equation}
Observe that this depends only on the IBO (equal over all antennas) and the number of transmit antennas. Fortunately, in this case by increasing the number of antennas, the SDR will be proportionally increased. 

However, if the distortion signal is correlated with the wireless channel, e.g., in LoS channel, the distortion can have the same spatial patter and gain as the wanted signal \cite{Larsson_2018_nonlinear_Los,Wachowiak2023}, i.e., $D^{\mathrm{RX}}$ will be $K$ times higher, resulting in SDR:
\begin{equation}
    SDR^{\mathrm{correlated}}=\frac{\lambda^2}{\frac{2}{3} \left(1-e^{-\gamma}-\lambda^2 \right)}, 
   \label{eq_SDR_correlated}
\end{equation}
that is independent of the number of utilized antennas.

Finally, an SDR can be calculated for a non-scheduled user. In this case, the channel $\tilde{h}_{n,k}$ will be uncorrelated with the precoder $w_{m,n,k}$, resulting in lack of the array gain $K$ in comparison to (\ref{eq_wanted_RX_power}). At the same time the distortion should be omnidirectionaly emitted from an array obtaining power as described by (\ref{eq_distortion_uncorrelated}). This results in 
\begin{equation}
    SDR^{\mathrm{victim}}=\frac{\lambda^2}{\frac{2}{3} \left(1-e^{-\gamma}-\lambda^2 \right)}, 
   \label{eq_SDR_victim}
\end{equation}
being equal to the $SDR^{\mathrm{correlated}}$ case. Observe that while this is observed at a victim UE, it can be treated as a linear interference to its nonlinear distortion power ratio.

While these equations define SDRs in specific cases, namely LoS and IID Rayleigh propagation channel, it is visible that the SDR can vary significantly, i.e., at least $K$ times.
The question is how significant is the PA nonlinearity problem in practical channels of partial inter-antenna or inter-subcarrier correlation, or when large-scale fading vary among antennas.

\section{3D Ray Tracer-based distortion power spatial distribution}\label{sec:ray-tracer-analysis}

The previous section introduced the system model and provided a theoretical analysis of SDR under the two state-of-the-art edge cases of radio channels, namely, uncorrelated Rayleigh and LoS radio channels. The aim of this section is to analyze the SDR, {\PK using (\ref{eq_SDR}),} under the close-to-real radio channels using the 3D-RT software. We also compare these results against the theoretical considerations. The main reason to follow the simulation studies rather than the analytical approach is that we avoid some simplifications, e.g., an analytical analysis for a soft-limiter model of PA would require its representation as a polynomial of limited order, decreasing the accuracy of the results. Moreover, our initial studies showed that direct simulation of the precoded OFDM signal going through the PA is less computationally complex than evaluation of the analytical model from~\cite{mollen_spatial_char_2018}. For the remaining part of the paper, we focus on a single UE case, i.e., MISO. This is the most challenging scenario, where potentially all the nonlinear distortion can be steered toward the UE~\cite{mollen_spatial_char_2018}. Moreover, unlike state-of-the-art works, e.g.,~\cite{bjornson2014massive}, we focus on the OFDM system, taking into account the intermodulation between the subcarriers caused by the nonlinear distortion.

\subsection{Simulation Setup} \label{subsec:simulation_setup}

\begin{figure}[!htb]
\centering
\includegraphics[trim={5cm 0.0cm 3cm 0cm}, clip,width=0.49\textwidth]{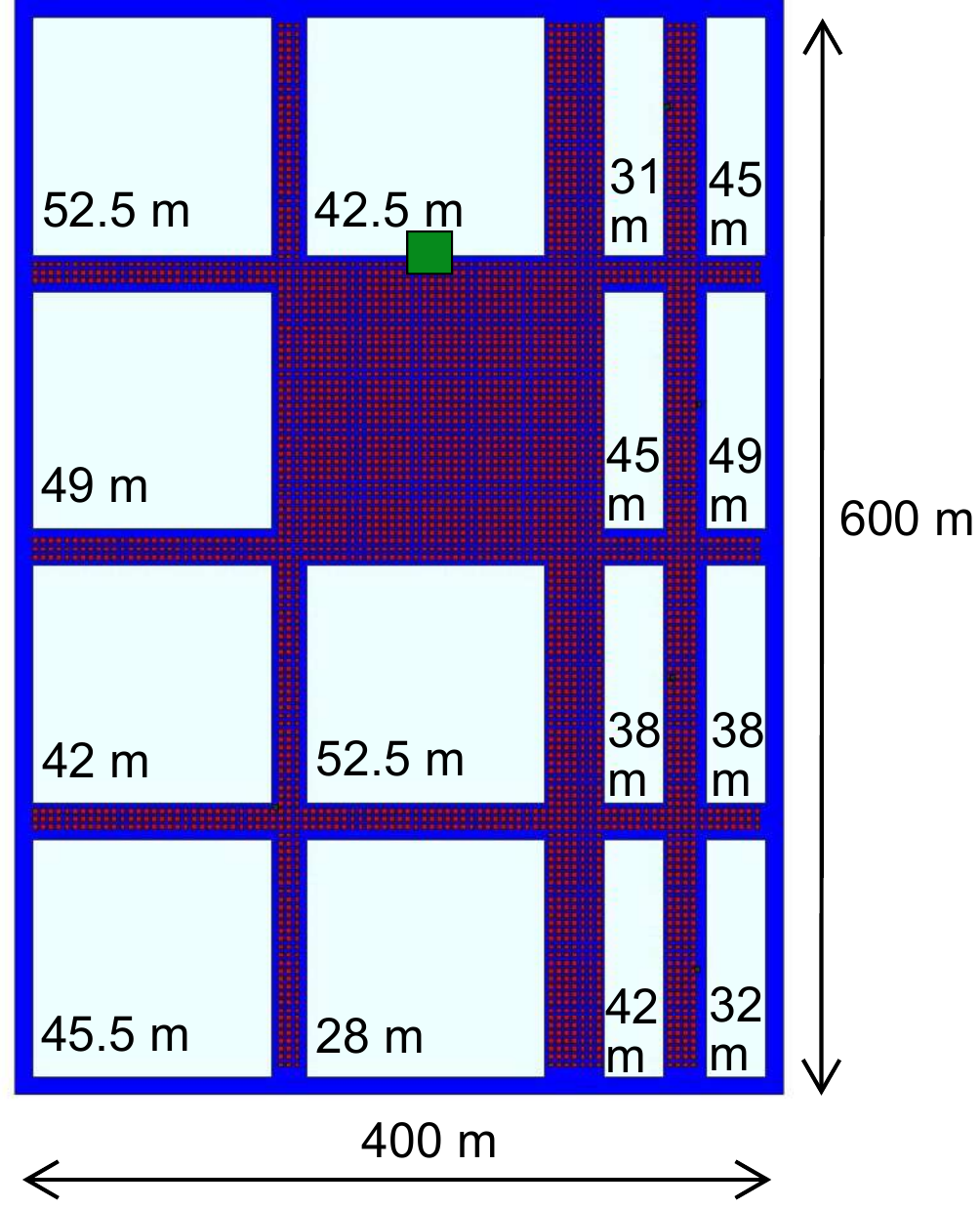}
\caption{3D Madrid Grid Model for the 3D-RT-based radio channel generation. The green and red rectangles are the MISO BS and the UEs, respectively. Numbers on the buildings indicate their height.}
\label{fig:scenario_model_2D}
\end{figure}

To evaluate the nonlinear effects in the downlink of the considered MISO system, under the realistic radio channel, we used Wireless InSite 3D-RT software. 3D-RT, tracks individual signal paths between each transmit and receive antenna, including reflections and diffractions from buildings. This results in a multi-path, frequency-selective fading model of the radio channel with spatial correlations between antennas being crucial for the M-MIMO system modeling. Using the Wireless InSite 3D-RT software, we first developed a 3D model of the Madrid Grid urban scenario proposed in the METIS project~\cite{metis2020}. 
Fig.~\ref{fig:scenario_model_2D} depicts the bird's-eye view of the built Madrid Grid model, together with its dimensions, i.e., width, length, and height of buildings. There is a MISO BS (marked with a green rectangle equipped with $K=128$ antennas arranged in a rectangular array ($8$ vertical $\times$ $16$ horizontal). It is installed at a height of $45$~m above ground level. Each antenna is associated with an independent transmitter chain that ends with a soft-limiter PA of $P_\mathrm{max,k}=2$~mW. Most importantly,  while the selected saturation power is arbitrary, the observed SDR depends on the IBO value, not the saturation power, as shown in Sec.~\ref{sec:system_model}. Therefore, observed results are universal. 
Through the analysis, a fixed IBO, averaged over frontends, will be used as defined in \eqref{eq:ibo}. As the power allocated to each PA is a result of MRT precoding according to \eqref{eq:mrt_precoder}, their operation point, i.e., $\gamma_k$ {\PK as defined in~\eqref{eq:ibo_k}}, may vary between transmitter chains. The BS operates at carrier frequency $f_c=3.6$~GHz, with $N_U=69$ 
subcarriers, and subcarrier spacing equal to the $360$~kHz (corresponds to the bandwidth of a single Resource Block (RB) in the 5G system with $12$ subcarriers per RB, and subcarrier spacing equal to the $30$~kHz). For each subcarrier independently, a precoder is calculated. Then, based on the precoder, we transmit 100 randomly modulated OFDM symbols using Quadrature Phase Shift Keying (QPSK) modulation.

To evaluate the nonlinear effects under various channels, we have defined a square grid of UEs with spatial resolution equal to 4~m, visible in Fig.~\ref{fig:scenario_model_2D} as red rectangles. There are 3542 UEs in total, each equipped with a single omnidirectional antenna placed at a height of $1.5$~m, as considered in 3GPP reference configurations~\cite{3gpp38901}. To obtain radio channel coefficients, we have configured the Wireless InSite software to consider 15 reflections and one diffraction. As a separate model, we have implemented the widely used (e.g.,~\cite{Guerreiro2016, Xu2024}) uncorrelated Rayleigh radio channel model for small-scale fading, while large-scale fading is modeled via the 3D-RT. In this case, the radio channel coefficient at the $n$-th subcarrier between the $k$-th BS antenna and the $m$-th single-antenna UE is a sample of an uncorrelated complex Gaussian random variable:
\begin{equation}
  h_{m,n,k} \sim \mathcal{C}\mathcal{N}(0, \beta_{m}),  
\end{equation}
where $\beta_{m}$ stands for the channel gain between the BS and $m$-th UE obtained from the 3D-RT. The simulation parameters are summarized in Tab.~\ref{tab:scenario_parameters}. 
 \begin{table}[!t]
\renewcommand{\arraystretch}{1.3}
\caption{Simulation Parameters}
\label{tab:scenario_parameters}
\centering
\begin{tabular}{|c|c|}
\hline
Parameter & Value \\
\hline
number of antennas $K$& $128$ ($8$ vertical $\times$ $16$ horizontal) \\
BS antenna height &$45$~m \\
PA max output power $P_{\mathrm{max},k}$ &$2$~mW\\
carrier frequency $f_c$ &$3.6$~GHz \\
number of used subcarriers $N_U$ &$69$ \\
subcarrier spacing  &$360$~kHz\\
Number of OFDM symbols& 100 \\
Modulation& QPSK \\
UE antenna & single omnidirectional \\
UE antenna height  &$1.5$~m \\
number of UEs &3542 \\
UE deployment & square grid of $4$~m resolution \\
3D-RT configuration & $15$ reflections, $1$ diffraction \\
\hline
\end{tabular}
\end{table}
The utilized Madrid Grid urban scenario provides both Close-to-Line-of-Sight (Close-to-LoS), as there is still multipath propagation expected, and Non-LoS (NLoS) propagation conditions. The former occurs, e.g., in the square placed in the middle, while the latter occurs for the UEs placed in the narrow street canyons. 

\subsection{Signal to Distortion Analysis for Victim UEs} \label{subsec:sdr_victim}
\begin{figure}[!htb]
\centering
\includegraphics[trim={0cm 0cm 19cm 1cm}, clip,width=0.49\textwidth]{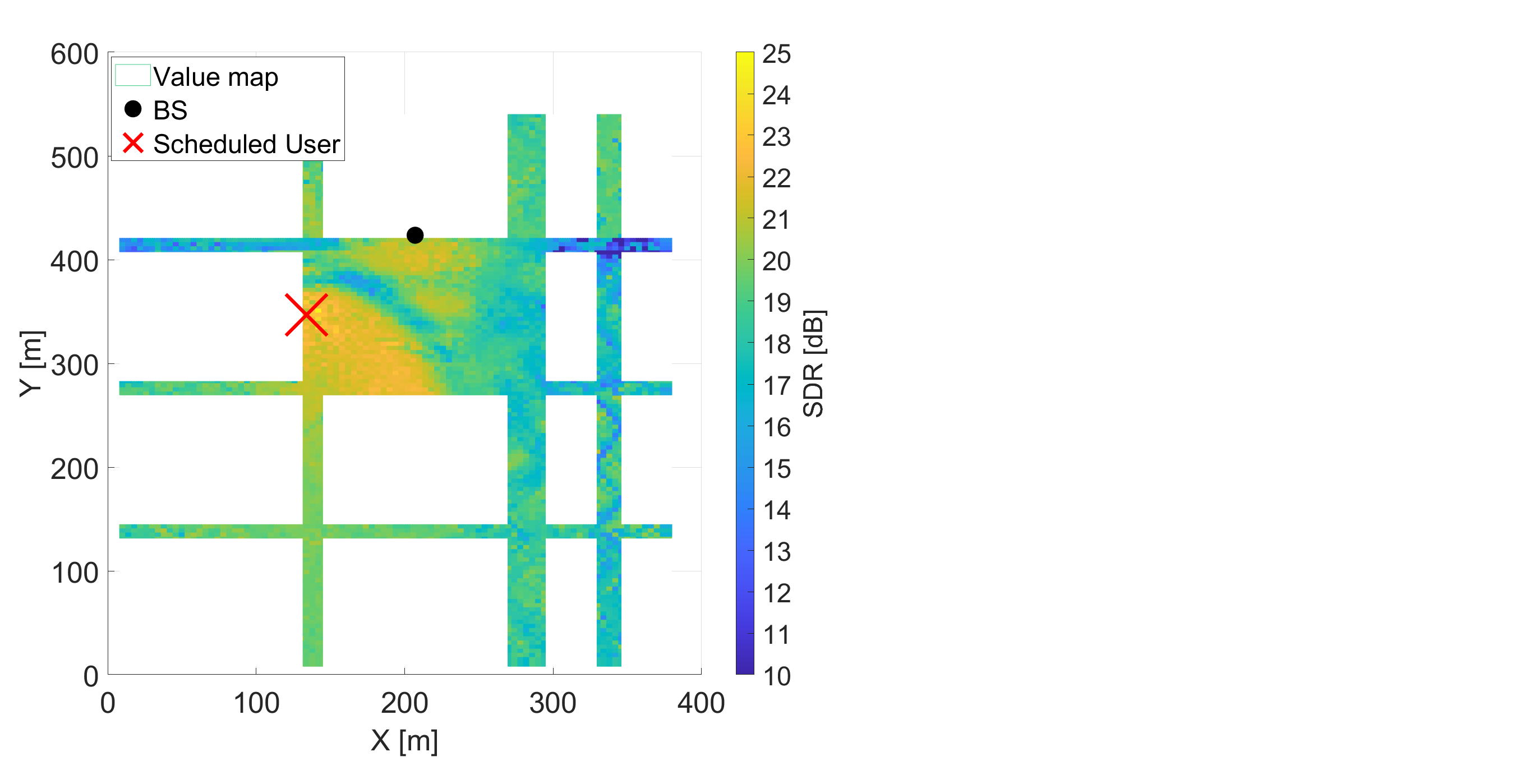}
\caption{SDR for \emph{victim UEs} under 3D-RT radio channel, while the scheduled UE (red cross) is under Close-to-LoS conditions. The {\MH IBO $\gamma = 3$~dB.}} 
\label{fig:lin2dist_ibo3_los}
\end{figure}
\begin{figure}[!htb]
\centering
\includegraphics[trim={0cm 0cm 19cm 1cm}, clip,width=0.49\textwidth]{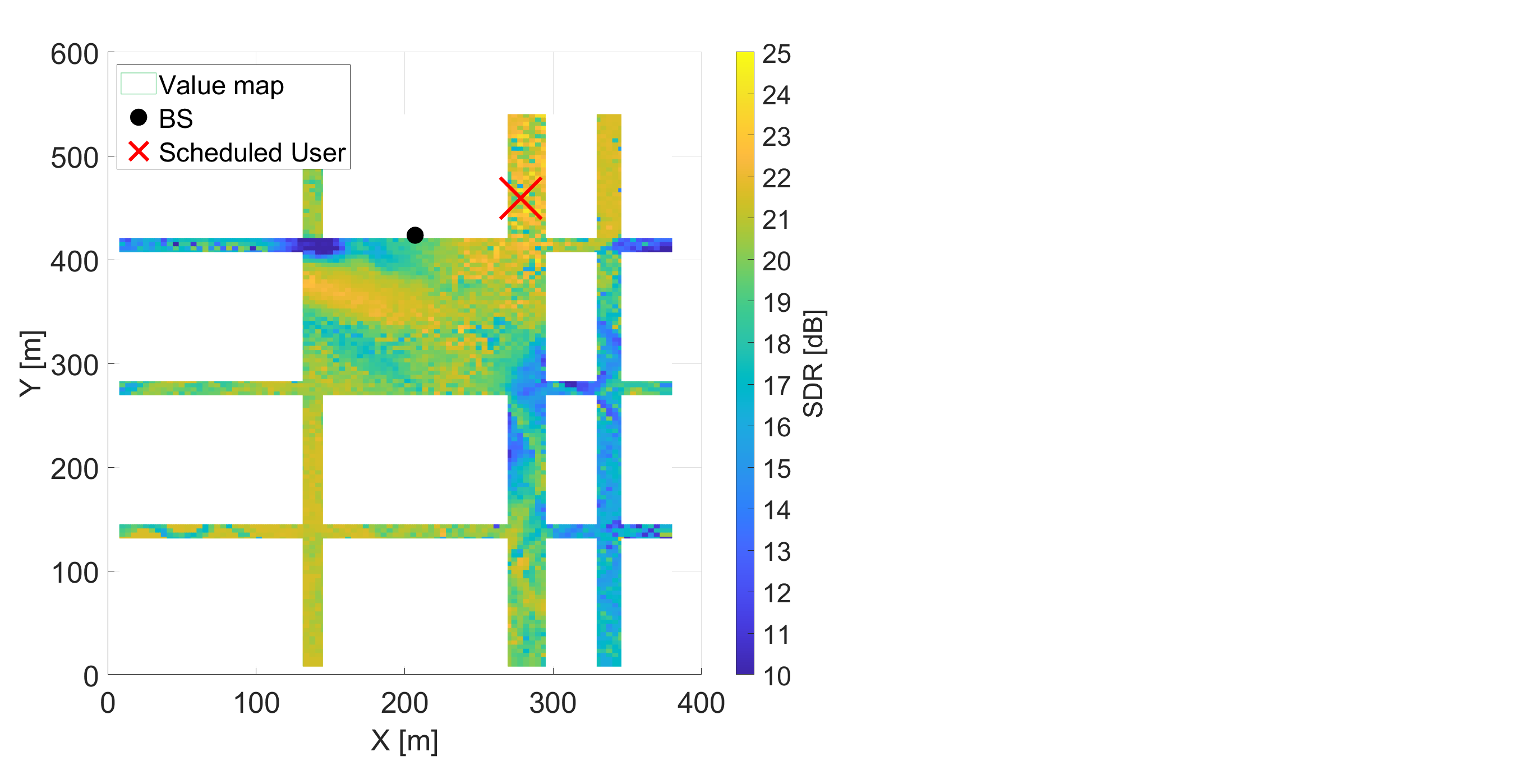}
\caption{SDR for \emph{victim UEs} under 3D-RT radio channel, while the scheduled UE (red cross) is under NLoS conditions. The {\MH IBO $\gamma = 3$~dB.}}
\label{fig:lin2dist_ibo3_nlos}
\end{figure}
Although in the considered MISO scenario there is no inter-user interference, as there is only one UE scheduled, in a real-world scenario, there can exist some UEs connected to other cells operating at the same frequency band. From this perspective, it is beneficial to investigate the SDR for the \emph{victim UEs}, i.e., UEs that are not currently scheduled but can be potentially served by the neighboring cells.
We can also treat these UEs as victims of Electromagnetic Field (EMF) exposure from the M-MIMO BS. While this is typically modeled assuming linear PA, the existence of nonlinear distortion can be an important factor in increasing EMF exposure\cite{Buzzi_2024_EMF_MMIMO}.
The SDR for \emph{victim UEs} under 3D-RT radio channel, while the scheduled UE (red cross) is under Close-to-LoS conditions and $IBO=3$~dB, is depicted in Fig.~\ref{fig:lin2dist_ibo3_los}. In some of the works,e.g.,~\cite{bjornson2014massive}, the nonlinear distortion is claimed to be uncorrelated with the wanted signal. This is true for the IID Rayleigh radio channel, where, according to \eqref{eq_SDR_victim}, the SDR of \emph{victim UEs} equals $19$~dB
for IBO of 3 dB. Observe that this equation assumes that both the wanted signal and the distortion do not add coherently at the reception point. 
However, in the case of realistic 3D-RT radio channels, a strong spatial correlation can be visible. The SDR in Fig.~\ref{fig:lin2dist_ibo3_los} is relatively high, i.e., about 23~dB, in many areas, including the area around the scheduled UE (red cross). This can be explained by coherent combining of the wanted signal (increasing the numerator of SDR), while the nonlinear distortion adds non-coherently. But in other locations, it drops to a level below 10~dB in the most extreme case. This can be attributed to, e.g., unequal IBO per antenna, as will be explained in detail later.

In Fig.~\ref{fig:lin2dist_ibo3_nlos}, a similar analysis is presented for a single scheduled NLoS UE. While a different spatial distribution of SDR is visible in the 3D-RT 
channel, the main conclusions can be the same as in the Close-to-LoS channel. High spatial variability of SDR is observed, spanning many dB above and below the theoretical value of $19$~dB. The SDR for \emph{victim UEs} depends on the precoder of a scheduled UE and radio channel coefficients, following non-trivial spatial patterns. 

\begin{figure}[!t]
\centering
\includegraphics[trim={0.0cm 0cm 0cm 0cm}, clip,width=0.49\textwidth]{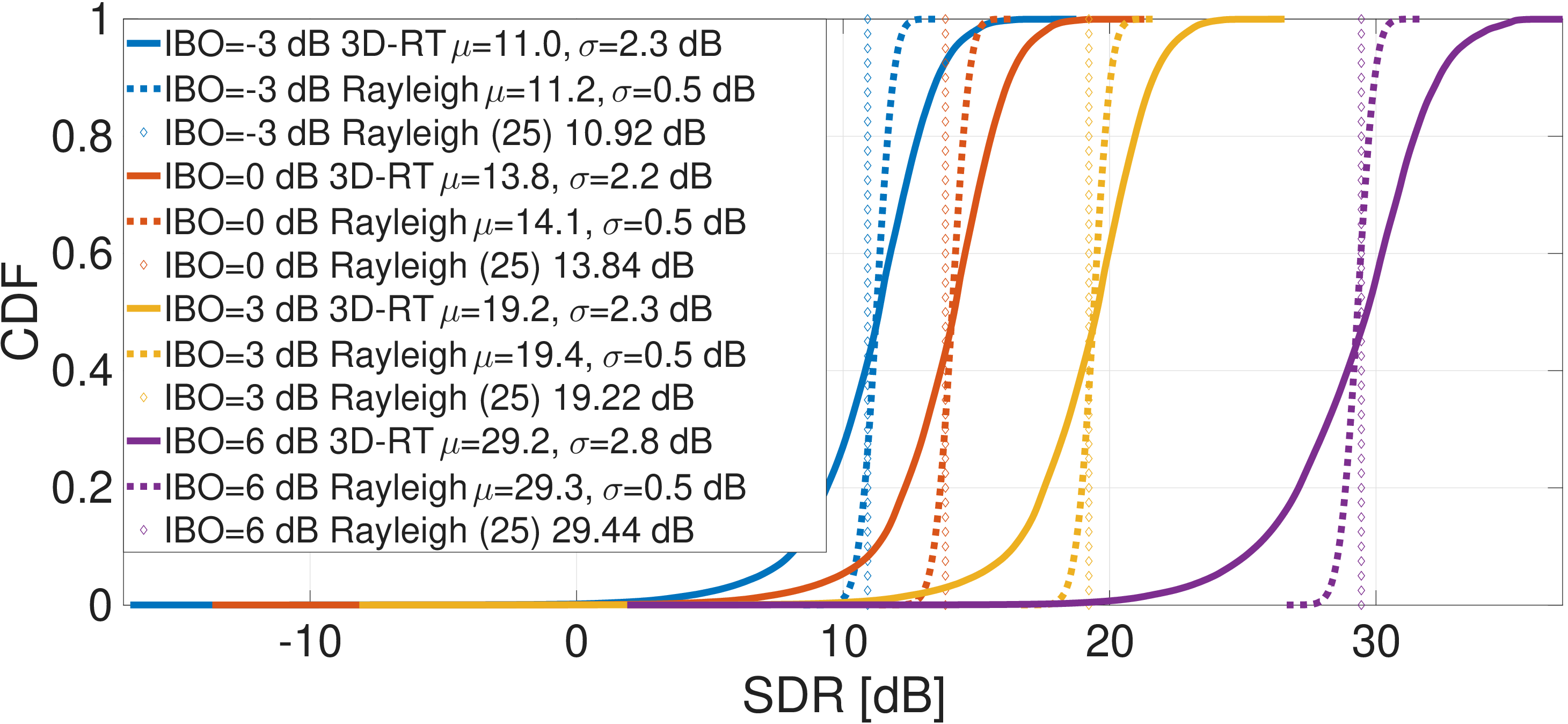}
\caption{ Comparison of SDR Cumulative Density Function (CDF) for \emph{victim UEs} under 3D-RT (simulation) and Rayleigh (simulation and theoretical - \eqref{eq_SDR_victim}) radio channel, for a dataset of 120 scheduled UEs with {\MH IBO $\gamma=\{-3, 0, 3, 6 \}$~dB.} 
}
\vspace{0.25cm}
\label{fig:lin2dist_cdf}
\end{figure}
\begin{figure}[!t]
\centering
\includegraphics[width=0.49\textwidth]{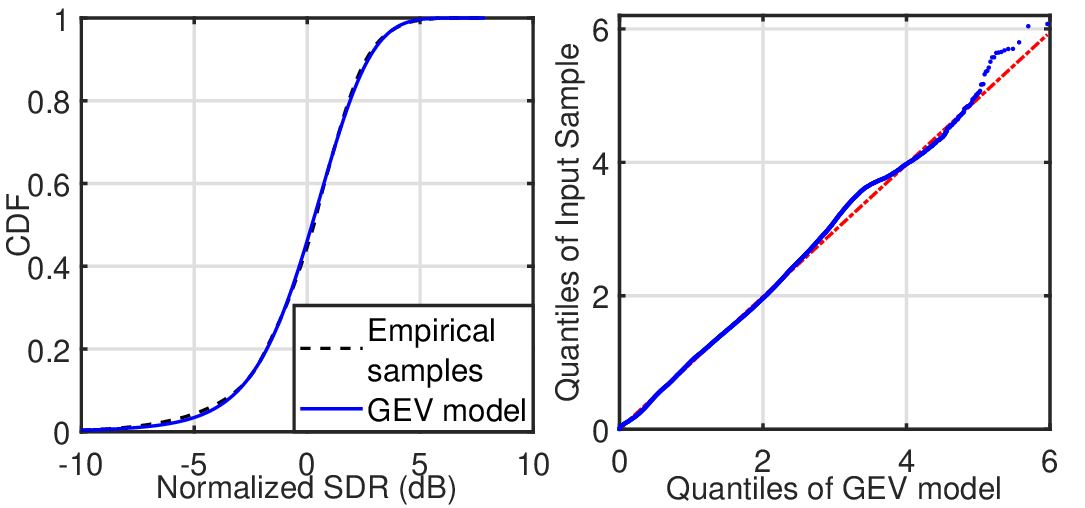}
\caption{ {\PK CDFs and Quantile-Quantile plots of normalized SDR values and their GEV distribution-based model}}
\vspace{0.25cm}
\label{fig:GEV_CDF}
\end{figure}

To provide better insights into the statistical characteristics of SDR for \emph{victim UEs} we have collected their SDR values, i.e., in 3542 \emph{victim UE} locations,  in each case of 120 scheduled UEs (placed on an equally spaced grid to reflect various LoS/NLoS radio conditions) under {\MH IBO $\gamma=\{-3, 0, 3, 6\}$}~dB, and both 3D-RT (simulation) and Rayleigh (simulation and theoretical results)  radio channels. We have estimated the distribution of \emph{victim UE} SDR as depicted in Fig.~\ref{fig:lin2dist_cdf}. 
For each plot, the mean value ($\mu$) and standard deviation ($\sigma$) are given. First, as expected based on SISO OFDM analysis\cite{kryszkiewicz2023efficiency}, mean SDR rises with IBO. Secondly, the mean value of SDR for 3D-RT is similar to the mean value for the simulated Rayleigh channel, and both follow the theoretical value of (\ref{eq_SDR_victim}). However, the difference is in the standard deviation.
It can be seen that the distribution for the simulated Rayleigh channel is almost a constant value with a negligible standard deviation from the theoretical value equal to about 0.5~dB. This result is probably due to a finite number of antennas and the number of subcarriers used for SDR estimation in each location, using one instance of a Rayleigh fading channel. On the other hand, in the case of the 3D-RT radio channel, the standard deviations are higher than $2$~dB. Distribution fitting was performed based on all measurements for all IBO values. The closest match (with SDR difference smaller than 0.4 dB for any CDF percentile) is obtained for the GEV distribution applied in the linear domain, allowing the estimated victim SDR to be defined as 
{\PK
\begin{equation}
    \tilde{SDR}^{\mathrm{victim}}\!\!\!\!={SDR}^{\mathrm{victim}} \cdot  GEV\left(\mu,\sigma,\xi \right)
   \label{eq_SDR_victim_model}
\end{equation}
where ${SDR}^{\mathrm{victim}}$ is obtained from (\ref{eq_SDR_victim}) using IBO value, and $GEV(\mu,\sigma,\xi)$ denotes truncated, for negative values, GEV random variable of location $\mu$, scale $\sigma$, and shape $\xi$. The 95\% confidence intervals for the estimated parameters are $\mu\in(0.8803,0.8817)$, $\sigma\in(0.4581,0.4592)$, and $\xi\in(-0.0447,-0.0429)$. The goodness of fit can be discussed. First, in Fig. \ref{fig:GEV_CDF} on the left, the CDF of the SDR normalized by the values of $SDR^{\mathrm{victim}}$, along with the CDF of the proposed GEV model is shown. Both CDFs are visibly well aligned. This is confirmed by the QQ plot on the right, showing a good match between both distribution quintiles. Moreover, the Kolmogorov-Smirnov (KS) test was performed, confirming that both distributions can be the same with a significance level of 0.01. However, a remark is needed as the KS test assumes independent data \cite{zeimbekakis2024misuses}. In our case, the SDR values are spatially correlated as visible, e.g., in Fig. \ref{fig:lin2dist_ibo3_los}. The correlation will be quantified in the next part of this section. While the SDR samples are collected on a rectangular, 4-meter grid, the data have to be preprocessed before a KS test to use only uncorrelated samples. This is done by downsampling the SDR set by picking only a single sample from every 200. This resulted in the KS test using 9643 distribution samples, which is still a high number, allowing statistically significant conclusions to be drawn}.    
Observe that the GEV random variable provides some scaling of the theoretical victim SDR value. This can be used for modeling stochastically the power of nonlinear distortion arriving at \emph{victim UE}.   

\begin{figure}[!t]
\centering
\includegraphics[trim={2cm 0cm 3cm 0cm}, clip,width=0.49\textwidth]{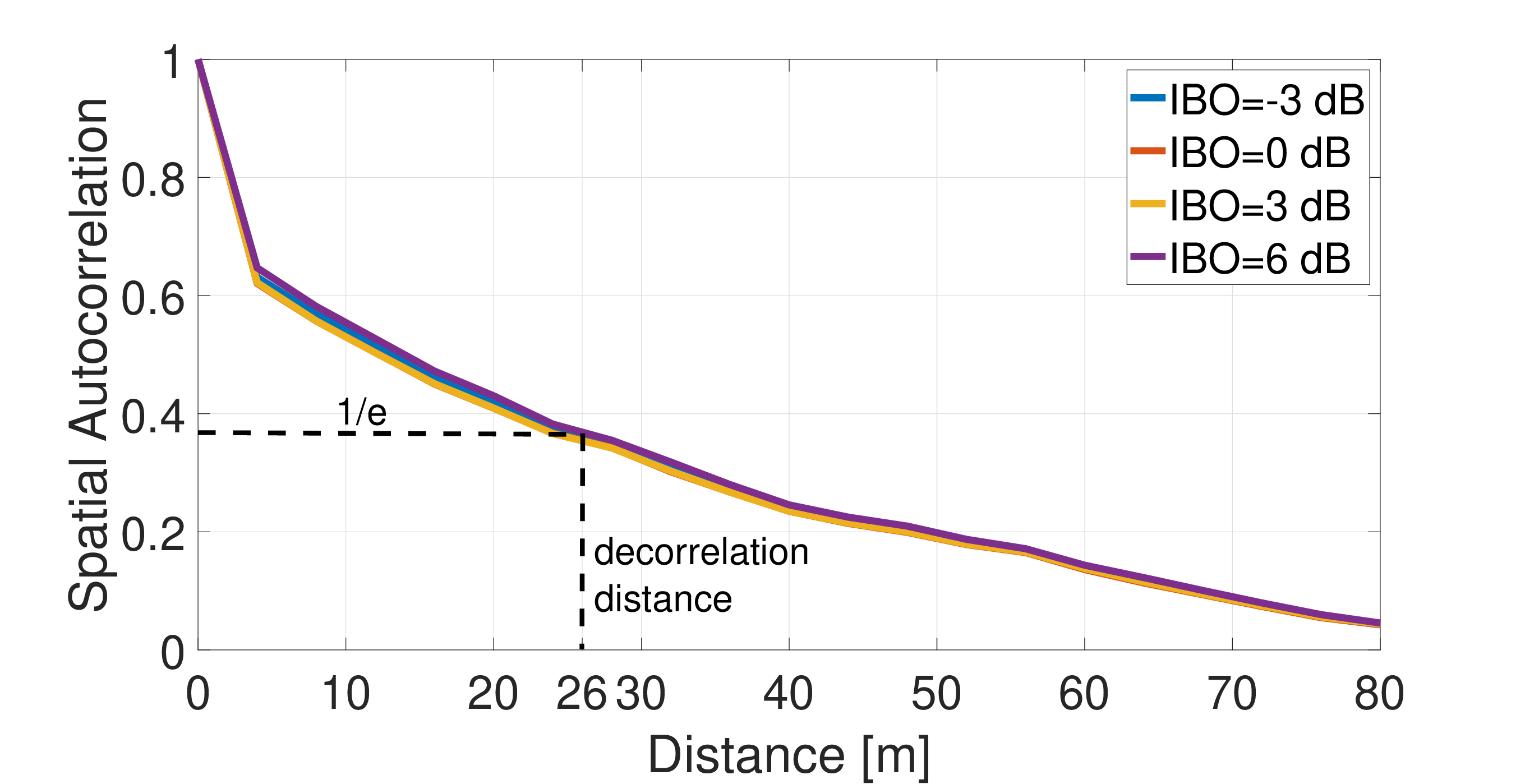}
\caption{ Spatial autocorrelation of SDR for \emph{victim UEs} under 3D-RT radio channel. The results are averaged over a scheduled 120 UE with {\MH IBO $\gamma=\{-3, 0, 3, 6 \}$~dB.}}
\vspace{0.25cm}
\label{fig:victim_spatial_correlation}
\end{figure}

Additionally, observe that both in Fig.~\ref{fig:lin2dist_ibo3_los} and Fig.~\ref{fig:lin2dist_ibo3_nlos}  local correlation of SDR values is observed, e.g., \emph{orange} points are typically surrounded by \emph{orange} points. Therefore, using the same SDR data as above, for 120 scheduled UEs, a spatial autocorrelation of the \emph{victim UEs} SDR is calculated over vertical and horizontal directions. The results averaged over 120 scheduled UEs are depicted in Fig.~\ref{fig:victim_spatial_correlation}.
It can be seen that the spatial autocorrelation of SDR for \emph{victim UEs} does not significantly depend on the IBO. In the context of mobile networks, e.g., spatial correlation of the large-scale fading~\cite{zhang2012cor}, the correlation distance is the distance where the autocorrelation function reaches $\frac{1}{e}$. The results show that the decorrelation distance of \emph{victim UEs} SDR is equal to 26~m. This means that within 26~m, similar values of SDR can be expected. One should notice that further studies are required to evaluate spatial correlation at distances of 0 to 4~m, which is below the resolution of the simulated UE grid.

Considering the above observations, the prediction of SDR for \emph{victim UEs} seems to be a non-trivial task. Observe that in practice, the scheduled (interfering) UE can be assigned to a different BS. In such a case, radio channel coefficients between \emph{victim UEs} and interfering BS are not known, and SDR is hard to obtain using e.g., \cite{mollen_spatial_char_2018}. We suggest using the approximation proposed in~\eqref{eq_SDR_victim_model}, extended with the spatial decorrelation. A similar model is used for shadowing modeling in the Gudmundson model~\cite{Gudmundson1991}. 
The proposed model can be used for interference estimation during network planning, e.g., for an $IBO=6$~dB in 10 \% of worst cases, the SDR is below $25.5$~dB. This is about 5 dB degradation compared to the upper bound of SDR under the Rayleigh channel.

\begin{figure}[!t]
\centering
\includegraphics[trim={0.0cm 0cm 0cm 0cm}, clip,width=0.49\textwidth]{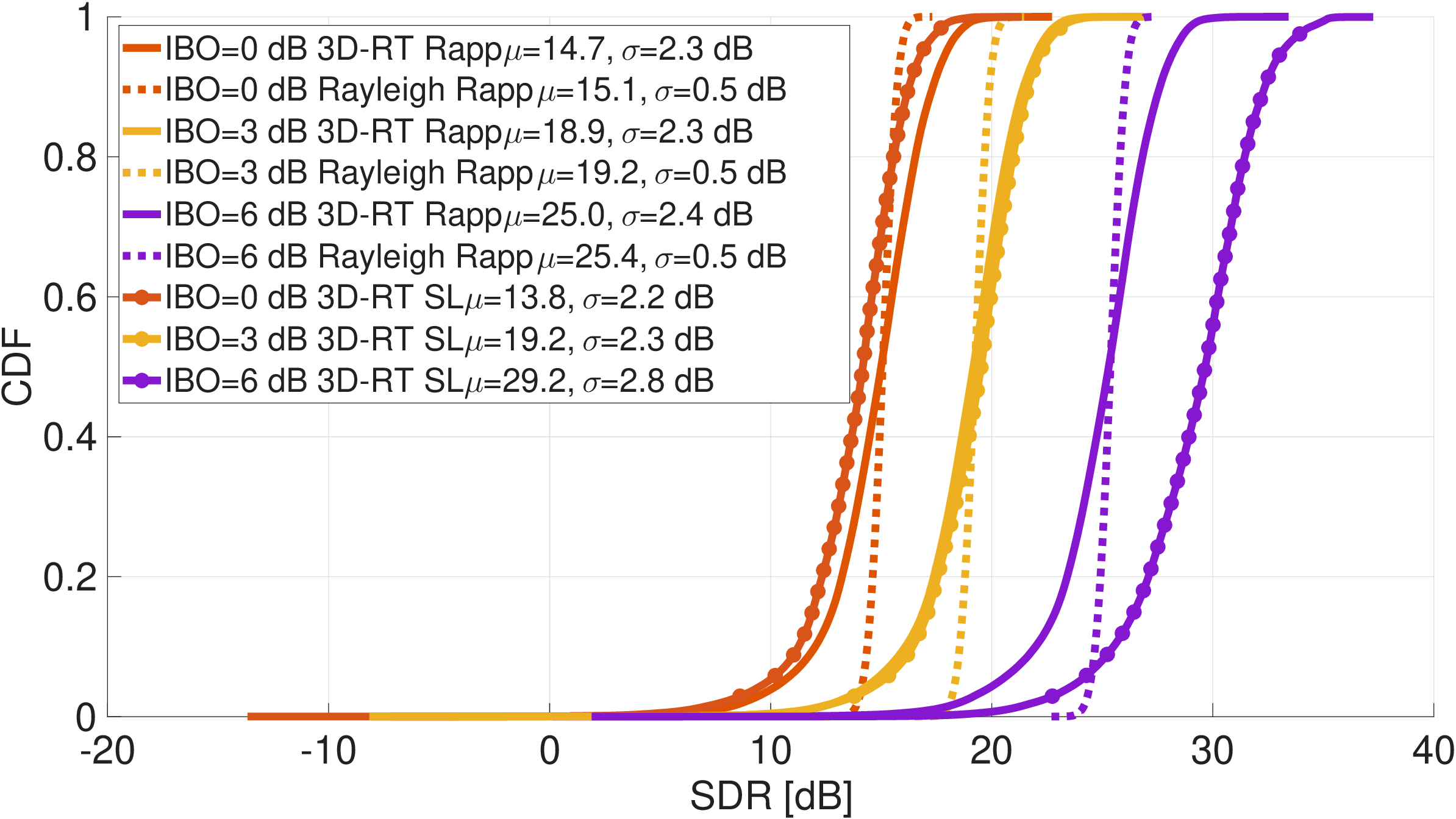}
\caption{ \MH Comparison of SDR distribution for \emph{victim UEs} under 3D-RT (simulation) and Rayleigh (simulation and theoretical - \eqref{eq_SDR_victim}) radio channel, for Rapp ($p=2$), and Soft-Limiter (SL) PA model and dataset of 120 scheduled UE with IBO $\gamma=\{0, 3, 6 \}$~dB. 
}
\vspace{0.25cm}
\label{fig:lin2dist_cdf_rapp2}
\end{figure}
{\MH As a last step, we extend our studies by replacing the soft-limiter PA model with the Rapp model~\eqref{eq:rapp_pa}, using a typical smoothing factor $p=2$~\cite{ochiai2013analysis}. The results are depicted in Fig.~\ref{fig:lin2dist_cdf_rapp2}, and compared against the previous ones obtained under soft-limiter. Under both Rapp and soft-limiter, the SDR under 3D-RT radio channel has a higher standard deviation and much longer distribution tails compared to the Rayleigh radio channel. It can be seen that under the 3D-RT radio channel, compared to the soft-limiter, Rapp PA is characterized by higher SDR for low IBO ($\gamma=0$ dB), and lower SDR for high IBO ($\gamma=6$ dB). However, this is a phenomenon related to the PA characteristics itself, unrelated to the multiple antennas, as is visible for the SISO system, e.g., by evaluating SDR using the methodology from \cite{kryszkiewicz2023efficiency}.
Most importantly, while the GEV SDR model is proposed for soft-limiter in~\eqref{eq_SDR_victim_model}, it also provides a good fit for Rapp PA, requiring only the ${SDR}^{\mathrm{victim}}$ to be calculated for the Rapp model, e.g., using numerical integration of formulas provided in \cite{kryszkiewicz2023efficiency}. This strengthens the hypothesis that the proposed GEV can be applied beyond the soft-limiter PA case.}


\subsection{Signal-to-Distortion Ratio Analysis for Scheduled UEs}
\begin{figure}[!htb]
\centering
\includegraphics[trim={0cm 0cm 22cm 1.5cm}, clip,width=0.49\textwidth]{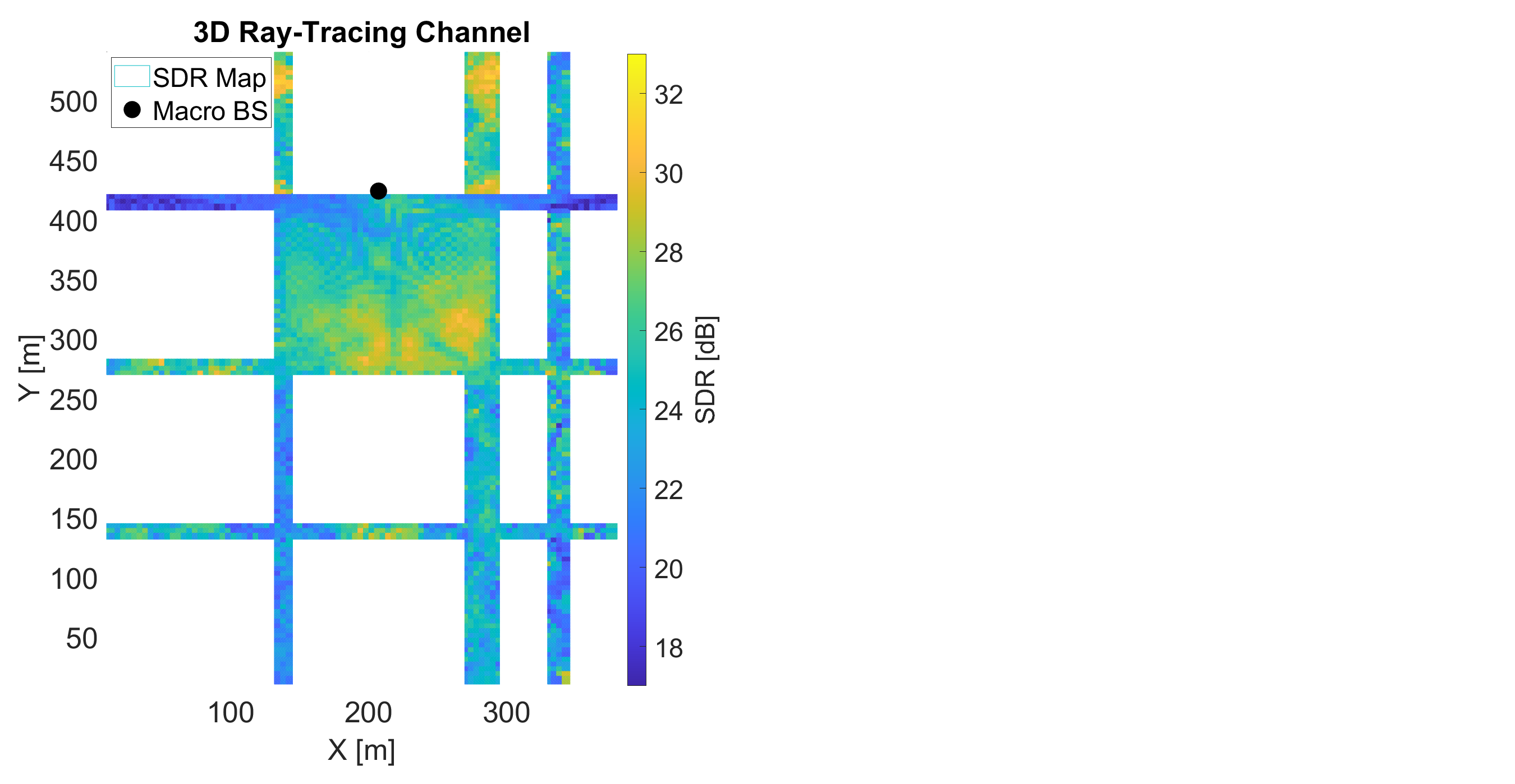}
\caption{Spatial distribution of SDR for \emph{scheduled UEs} 3D-RT radio channel, for {\MH IBO $\gamma = 3$~dB.}}
\label{fig:lin2dist_ibo3_scheduled}
\end{figure}

Within this subsection, the SDR of the currently scheduled UE is investigated under the same simulation setup as previously (see Tab.~\ref{tab:scenario_parameters}), i.e., 3542 positions of a single UE have been investigated. This SDR is important from the perspective of optimizing a given UE rate, e.g., by finding a proper transmission power. 
The spatial distribution of SDR for \emph{scheduled UEs} under 3D-RT radio channel, for {\MH IBO $\gamma = 3$~dB}, is depicted in Fig.~\ref{fig:lin2dist_ibo3_scheduled}. The results for a 3D-RT radio channel indicate a high spatial variability of the \emph{scheduled UEs} SDR. This can be caused by the specific radio channels that correspond to the particular location of a UE. This is in line with the analysis in \cite{mollen_spatial_char_2018,Bjorson_2023_nonlinearities} showing that the SDR in the LoS scenario can be much lower than in iid Rayleigh scenario, as shown theoretically by (\ref{eq_SDR_correlated}) and (\ref{eq_SDR_uncorrelated}), respectively. In the considered case of {\MH IBO $\gamma= 3$~dB}, the observed values range from about $18$~dB to slightly above $32$~dB. The higher values are not reaching the theoretical SDR for the Rayleigh radio channel of about $40$~dB obtained using (\ref{eq_SDR_uncorrelated}). Even with hundreds of propagation paths between transmitter and receiver generated by a high-end 3D-RT, the obtained channels are not close enough to the iid Rayleigh channel to result in omnidirectional emission of the nonlinear distortion signal.  
Moreover, in some locations, the SDR obtained through the 3D-RT simulations drops below the theoretical value for LoS of about $19$~dB, which will be attributed to unequal power per antenna in the later part of this section. Most interestingly, the visibility between the reception point and the BS location is not visibly correlated with the achievable SDR, e.g., there are points on the square in front of the BS with LoS visibility and SDR of $30$~dB, being closer to the iid Rayleigh channel case.  

\begin{figure}[!t]
\centering
\includegraphics[trim={0.0cm 0cm 0cm 0cm}, clip,width=0.49\textwidth]{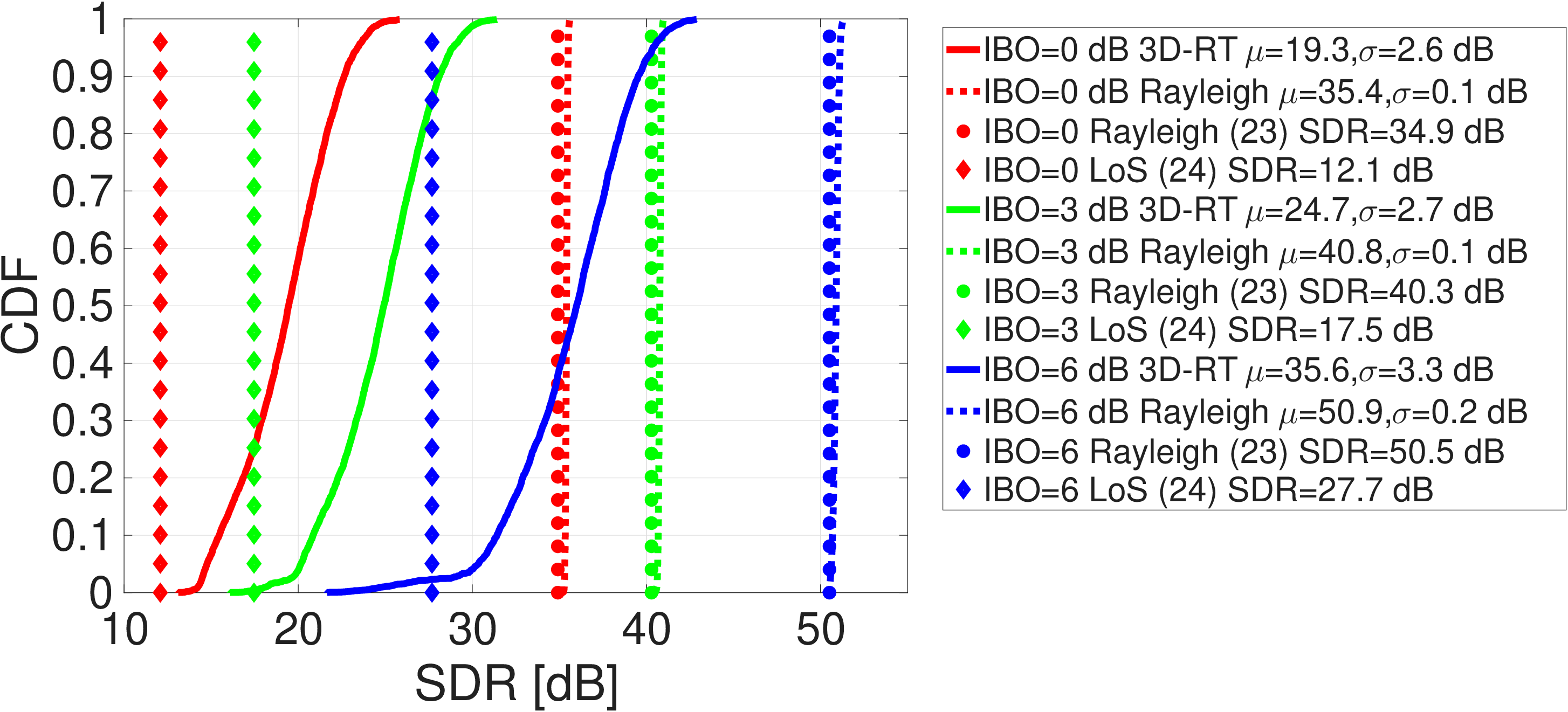}
\caption{Comparison of SDR distribution for scheduled UEs under 3D-RT radio channel against theoretical SDR for  Rayleigh~\eqref{eq_SDR_uncorrelated}, and LoS~\eqref{eq_SDR_correlated} radio channel, with {\MH IBO $\gamma=\{0, 3, 6 \}$~dB.} 
}
\label{fig:sdr_scheduled_cdf}
\end{figure}

To provide quantitative insights into the properties of SDR distribution for \emph{scheduled UEs} for various IBO values simulated under both Rayleigh and 3D-RT propagation channel models, the CDF plots are shown in Fig.~\ref{fig:sdr_scheduled_cdf}. They are supported by the plots of theoretical SDRs for uncorrelated and fully correlated distortions obtained using  \eqref{eq_SDR_uncorrelated} and \eqref{eq_SDR_correlated}, respectively. 
First, it can be seen that the theoretical approximation of SDR under the Rayleigh radio channel according to \eqref{eq_distortion_uncorrelated} fits well the simulation result, i.e., the difference between the simulated and approximated average SDR for the Rayleigh radio channel is about 0.5 dB. This difference can be caused, e.g., by a fixed $2/3$ constant as discussed above (\ref{eq_distortion_uncorrelated}). Even with only $100$ OFDM symbols used in simulations, the SDR is very stable, with the standard deviation of around 0.1 dB for iid Rayleigh channel. 

\begin{figure}[!t]
\centering
\includegraphics[trim={2cm 0cm 2cm 1cm}, clip,width=0.49\textwidth]{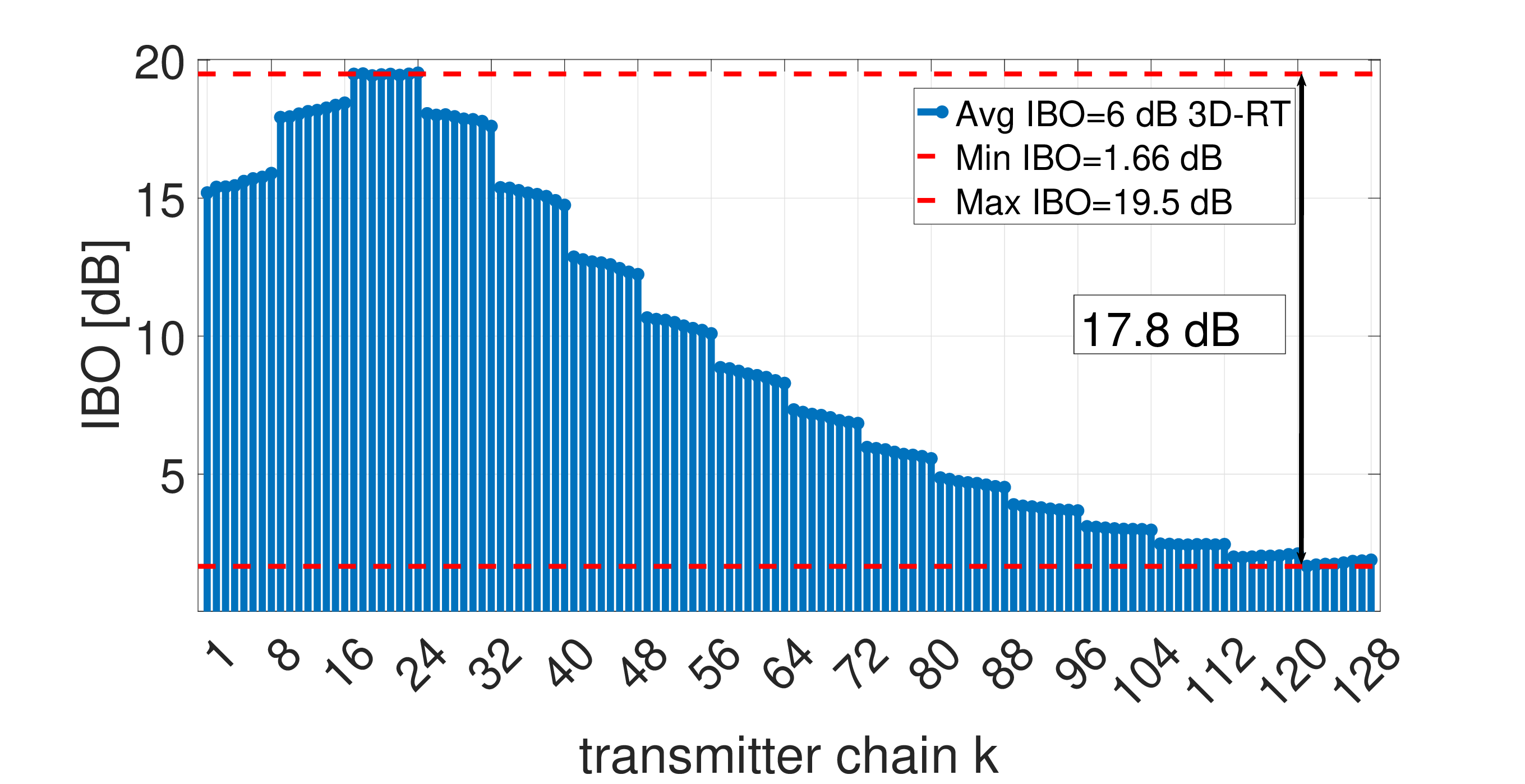}
\caption{IBO of $k$-th transmitter chain for a single UE (of the highest IBO variation) under average {\PK IBO $\gamma=6$~dB}.
}
\vspace{0.25cm}
\label{fig:per_antenna_sdr}
\end{figure}

As expected, the 3D RT radio channel results in significantly different SDR values than the two theoretical ones, with the range of possible SDR values spanning tens of dB for a fixed mean IBO value. Moreover, for each IBO there exists a small number of users whose signal quality is seriously distorted, below the typically assumed worst-case LoS result (\ref{eq_SDR_correlated}) \cite{mollen_spatial_char_2018}. In the LoS case, the same precoding gain is applied to both the wanted signal and distortion, preventing the M-MIMO system from increasing gain with the number of antennas, as discussed in Sec. \ref{sec:system_model}. The even further degradation of SDR visible on the CDF plots is caused by the fact that the MRT precoder can assign different amounts of power to certain transmitter chains, making their individual IBOs, i.e., $\gamma_k$ {\MH as defined in~\eqref{eq:ibo_k}}, significantly different. This can be possible, e.g., if some antennas observe a different electromagnetic shadow than others.  For instance, while the average IBO of all transmitter chains is about 6~dB, the IBO of individual PAs can vary by a few dB. An example of such IBO variation for a single UE out of the observation set, for which significant IBO variations are observed, is shown in Fig.~\ref{fig:per_antenna_sdr} for
 average {\MH IBO $\gamma=6$~dB.}
 In the figure, the red dashed lines mark the maximum and minimum observed values of IBO being 19.5 and 1.66~dB, respectively. In other words, although the average {\MH IBO $\gamma$ equals $6$~dB}, the values over individual PAs vary significantly. In this case, the nonlinear distortion power is dominated by components coming from frontends of the lowest IBO. This effect has not been discussed in previous works, e.g., \cite{mollen_spatial_char_2018}. The stairs-like characteristic of IBO in Fig.~\ref{fig:per_antenna_sdr} is related to the antenna array geometry, i.e., a rectangular array with $16$ rows and $8$ columns.

{\MH
\begin{figure}[!t]
\centering
\includegraphics[trim={0.0cm 0cm 0cm 0cm}, clip,width=0.49\textwidth]{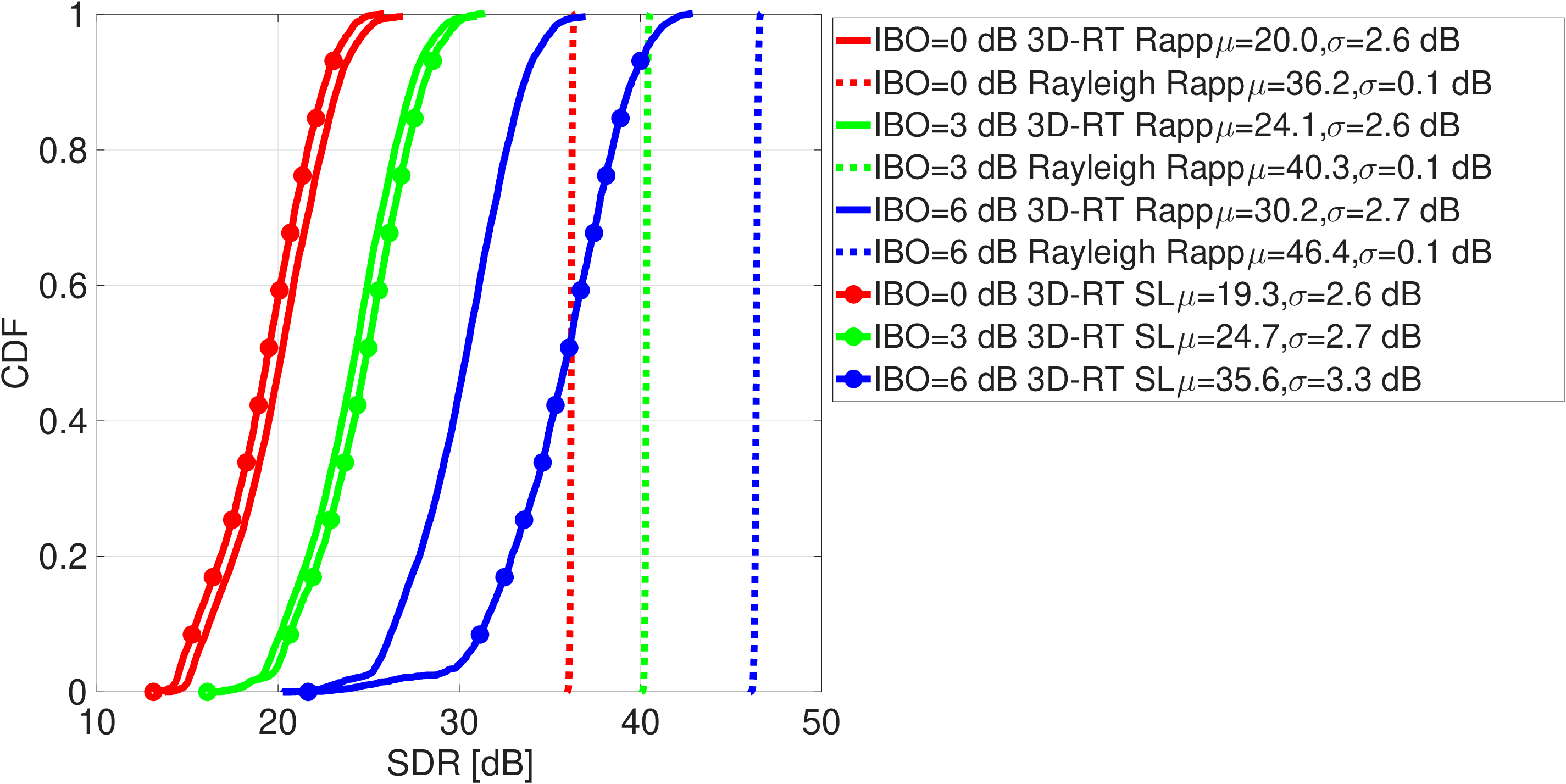}
\caption{{ \MH Comparison of SDR distribution for scheduled UEs for Rapp ($p=2$), and soft-limiter (SL), under 3D-RT and Rayleigh radio channel, with IBO equal $\gamma=\{0, 3, 6 \}$~dB.
}}
\label{fig:sdr_scheduled_cdf_rapp}
\end{figure}
}
{\MH Similarly to the analysis of SDR for \emph{victim UEs} in Sec.~\ref{subsec:sdr_victim}, we can extend the above analysis considering the Rapp PA. The comparison of SDR distributions under Rapp PA ($p=2$) and soft-limiter PA (SL) is shown in Fig.~\ref{fig:sdr_scheduled_cdf_rapp}. It can be seen that under the Rapp PA model, the SDR obtained under the 3D-RT radio channel model is lower compared to the Rayleigh radio channel, and follows a non-trivial distribution with a standard deviation of about $2.6$~dB, and long tails. This is similar to what was already observed and discussed for a soft-limiter. However, as expected from differences in PAs characteristics, there is a difference between SDRs under Rapp PA and soft-limiter in the 3D-RT channel. For low IBO, e.g., $\gamma=0$~dB, Rapp PA is characterized by higher SDR. while for higher IBO, e.g., $\gamma=6$~dB, the soft-limiter has much higher SDR. This results from the impact of different PA characteristics on the multicarrier system, even SISO one, and has already been revealed for \emph{victim UE}. See Fig.~\ref{fig:lin2dist_cdf_rapp2} and its discussion. Most importantly, this result shows that significant variations in SDR values are not only specific to the soft-limiter PA.}

The general conclusion from the analysis in this section is that to analyze the impact of nonlinear distortion on the MISO OFDM system, a proper and realistic radio channel model must be used. Under a much-simplified model like the uncorrelated Rayleigh radio channel model, the results might be misleading, resulting in too optimistic conclusions. Moreover, the achieved SDR depends not only on the average IBO but also on the distribution of the power between antennas. This distribution is a result of the MRT precoding allocating more power to antennas of higher channel gain to a given UE, causing SDR to fall even below the theoretical values for LoS, often considered as the worst-case scenario.
Most importantly, the SDR depends mostly on the observed radio channel coefficients. Fortunately, in the case of \emph{scheduled UEs}, the wireless channel can be estimated. While theoretically, for each channel impulse response, simulations can be carried out to obtain SDR, this will require significant time and computational complexity. Taking into account these limitations,
in Sec.~\ref{sec:ml_based_sdr_predioction} we will continue the analysis in order to propose an ML method aimed at the prediction of SDR based on the radio channel coefficients and IBO.   

\section{ML-based SDR Prediction} \label{sec:ml_based_sdr_predioction}

In the previous section, we have shown that there is a dependency between SDR and the per-antenna IBO. In extreme cases, such an IBO variation can lead to lower SDR than the one computed under LoS (being claimed as the worst-case scenario from the nonlinear perspective\cite{mollen_spatial_char_2018}) for the same average IBO. This is the result of the utilized MRT precoder based on the complex conjugate of the radio channel coefficients. From this perspective, we see that the SDR of the $m$-th UE depends on its radio channel. More specifically, the spatial correlation between antennas seems to have high influence on the observed SDR. This is visible via significantly different results in LoS and iid Rayleigh channel. The spatial characteristics of the M-MIMO channel are reflected by the correlation matrix~\cite{bjornson2017}. The correlation matrix of the $m$-th UE can be estimated using some independent channel realizations observed at the BS, e.g., for $N_U$ subcarriers in our case. The estimated correlation matrix for the $m$-th UE is given by:
\begin{equation}\label{eq:correlation_matrix}
    \hat{\mathbf{R}}_m = \frac{1}{N_U}\sum_{n=1}^{N_{U}} \mathbf{h}_{m,n}\mathbf{h}_{m,n}^H,
\end{equation}
where $\mathbf{h}_{m,n}=\{h_{m,n,k}\}_{k=1}^{K}$ is a vertical vector.
The diagonal of the correlation matrix contains real numbers and directly reflects the power distribution among antennas while utilizing MRT. The diagonal is directly related to IBO per antenna and, therefore, can be used as one input for ML-based SDR estimation. However, it is not sufficient.

We can show that the off-diagonal elements of this matrix will look different in extreme cases from the SDR value perspective, i.e., LoS and iid Rayleigh channel case.
First, as an example, let us now investigate the properties of the correlation matrix of the rectangular antenna array considered in Sec.~\ref{sec:ray-tracer-analysis} under the LoS radio channel. Under the assumption that the signal bandwidth is small with respect to the carrier frequency, the radio channel coefficients for the $m$-th UE are given by~\cite{planararrays2002}:
\begin{equation}\label{eq:channel_los}
    h_{m,n,k}^\mathrm{los}=\sqrt{\beta_m}e^{j\phi_{k}(\psi_m, \theta_m)},
\end{equation}
 where $\beta_{m}$ stands for the average channel gain between the BS and $m$-th UE, and $\phi_{k}(\psi_m, \theta_m)$ denotes phase shift at $k$-th antenna for the communications with UE located at azimuth angle $\psi_m$ and elevation angle $\theta_m$ from the BS. 
Let us denote by $l_k$ the horizontal distance, normalized by the wavelength, between the first antenna element and element $k$ over the array plane. Similarly, $r_k$ denotes the distance in the vertical plane.  
In such a case, the signal phase change for $k$-th antenna element equals~\cite{planararrays2002}: 

\begin{equation}
    \phi_{k}(\psi_m, \theta_m) \!=\! \pi \left(l_k\sin{\theta_m}\cos{\psi_m}\!+\!r_k\sin{\theta_m}\sin{\psi_m} \right). 
\end{equation}
As a result the element $\hat{R}_{m,k,\hat{k}}^\mathrm{los}$ of the correlation matrix $\hat{\mathbf{R}}_m^\mathrm{los}$, that describes correlation between $k$-th and $\hat{k}$-th antenna under LoS radio channel is given by:
\begin{equation} \label{eq:corr_mat_los}
\hat{R}_{m,k,\hat{k}}^\mathrm{los} = \beta_m e^{j\left(\phi_{k}(\psi_m, \theta_m)-\phi_{\hat{k}}(\psi_m, \theta_m)\right)}.    
\end{equation}
For the diagonal elements, $k=\hat{k}$ and phases vanishes resulting in $\hat{R}_{m,k,k}=\beta_m$. For the off-diagonal elements, $k \neq\hat{k}$, the magnitude of the complex cross antenna correlation is a fixed value:
\begin{equation}
    |\hat{R}_{m,k,\hat{k}}^\mathrm{los}| = \beta_m,
\end{equation}
while the phase (argument) depends only on the array geometry ($l_k$, and $r_k$) and $\psi_m$, $\theta_m$ angles:
\begin{align}
    \arg &\left\{
    \hat{R}_{m,k,\hat{k}} \right\} =
    \\ \nonumber
  \pi &\left\{ 
  \left[l_k-l_{\hat{k}} \right]\sin{\theta_m}\cos{\psi_m}+\left[r_k-r_{\hat{k}}\right]\sin{\theta_m}\sin{\psi_m} 
  \right\}.
\end{align}
It is important to note that the SDR for LoS (see \eqref{eq_SDR_correlated}) does not depend on the $\psi_m$, $\theta_m$ angles, but only on the IBO value. As such, the phase of $\hat{R}_{m,k,\hat{k}}$ seems to be obsolete to estimate SDR in this case. 

Let us now analyze the correlation matrix under the iid Rayleigh radio channel, being the other extreme for SDR value under a given $IBO$. According to \cite{bjornson2017}, the correlation matrix is given by:
\begin{equation}\label{eq:corr_mat_rayleigh}
    \hat{\mathbf{R}}_n^\mathrm{rayleigh} = \beta_m\mathbf{I}_K,
\end{equation}
where $\mathbf{I}_K$ is an identity matrix of size $K\times K$. 

We can see that in both cases the diagonal elements are the same, but the magnitude of cross-antenna correlation varies from $0$ in the case of the Rayleigh channel to $\beta_m$ in the case of LoS. As the SDR does not depend on the phase of $\hat{\mathbf{R}}_m^\mathrm{los}$ in case of the LoS channel, and in the case of the Rayleigh channel, off-diagonal elements of $\hat{\mathbf{R}}_m^\mathrm{rayleigh}$  are zeros we expect SDR to depend only on the $|\hat{\mathbf{R}}_m|$. Moreover, as both the desired signal and distortion go through the same radio channel, the SDR does not depend on the large-scale channel gain, thus $|\hat{\mathbf{R}}_m|$ can be normalized by the average channel gain $\beta_m$. 
After this normalization, the diagonal of $\frac{|\hat{\mathbf{R}}_m|}{\beta_m}$ reflects the differences between the average channel gain between antennas. If all channels have the same large-scale gains, all diagonal elements are equal to~$1$, otherwise each diagonal element is a scaling factor reflecting the per-antenna power allocation, i.e., fractor of the mean per-antenna power $1/K \sum p_k $. However, we would like the matrix to be related to the operating point of each PA that depends on the mean saturation power, i.e., $1/K \sum P_{\mathrm{max},k} $. While both $1/K \sum p_k $ and $1/K \sum P_{\mathrm{max},k}$ are included in IBO definition, we can divide $\frac{|\hat{\mathbf{R}}_m|}{\beta_m}$ by {\MH average IBO $\gamma$ \eqref{eq:ibo}} resulting in $\frac{1}{\gamma_k}$ on the diagonal, being related to SDR as shown in Sec.~\ref{sec:ray-tracer-analysis}. The above considerations result in the following definition of a feature matrix $\mathbf{F}_m$, taking into account both the correlation matrix, average channel gain, and IBO: 
\begin{equation}\label{eq:feature_matrix}
\mathbf{F}_m = \frac{1}{N_{U}\cdot \gamma \cdot \beta_m}\left| \sum_{n=1}^{N_{U}} \mathbf{h}_{m,n}\mathbf{h}_{m,n}^H \right|.    
\end{equation}
First, the feature matrix $\mathbf{F}_m$ represents on its diagonal the distribution of power between transceiver chains with respect to the saturation power of PAs. The off-diagonal elements represent the magnitude of correlation, reflecting whether the radio channel is closer to LoS (full correlation) or a Rayleigh (full decorrelation) radio channel.

\begin{figure*}[!t]
\centering
\includegraphics[trim={5cm 0cm 0cm 0cm}, clip,width=0.8\textwidth]{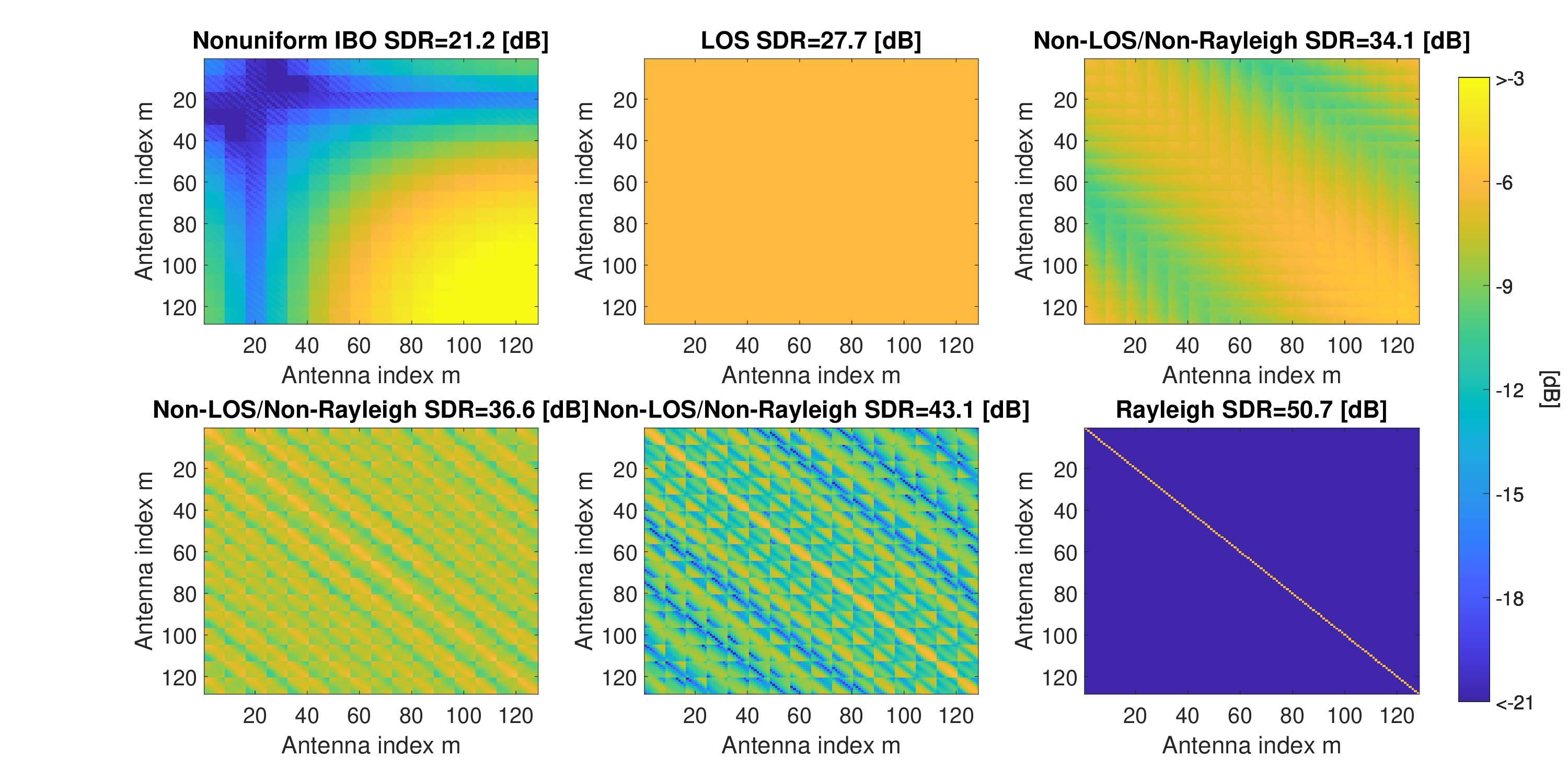}
\caption{Feature matrices representing various spatial properties of the radio channel.
}
\vspace{0.25cm}
\label{fig:feature_matrix}
\end{figure*}

The representative example of feature matrices obtained using the simulation framework from Sec.~\ref{sec:ray-tracer-analysis} is depicted in Fig.~\ref{fig:feature_matrix} for~$IBO=6$~dB. These are selected as the most distinctive, based on the SDRs from Fig.~\ref{fig:sdr_scheduled_cdf}.
The top left plot illustrates the worst observed SDR case of 21.2~dB, where significant non-uniform power distribution between transceiver chains is observed. This is over 6~dB below the SDR of LoS for uniform power allocation obtained using (\ref{eq_SDR_correlated}). The middle top picture illustrates the theoretical feature matrix for the LoS radio channel, while the bottom right picture stands for the theoretical feature matrix of the Rayleigh radio channel that is associated with the highest 
SDR equal to 50.7~dB. The remaining pictures are the feature matrices resulting from the 3D-RT simulations that cannot be categorized as either the LoS or the Rayleigh radio channel. 

\begin{figure}[!t]
\centering
\includegraphics[trim={0cm 1.5cm 0cm 1cm}, clip,width=0.49\textwidth]{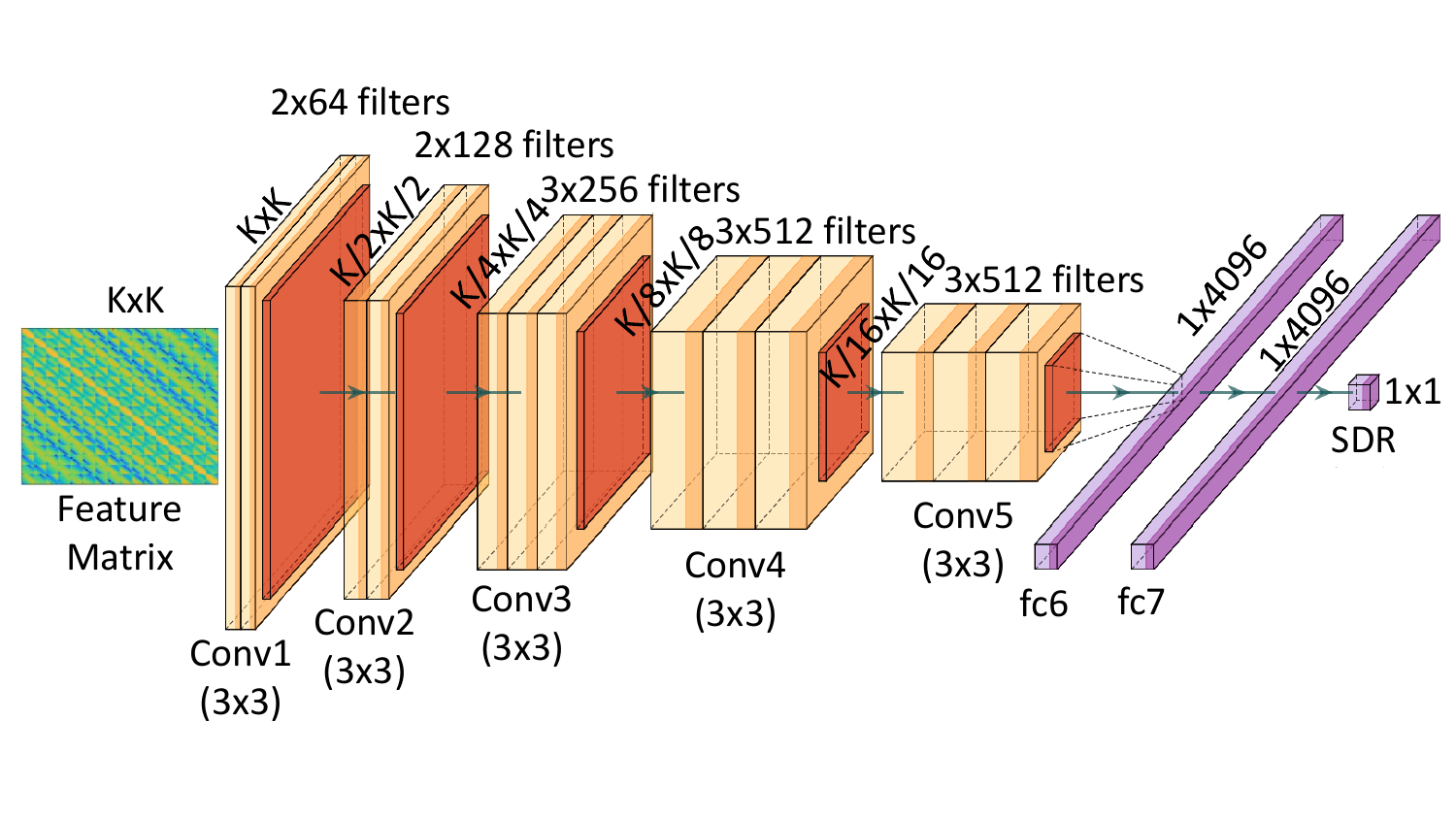}
\caption{Architecture of VGG16 CNN for SDR prediction. 
}
\vspace{0.25cm}
\label{fig:vgg16}
\end{figure}

The main hypothesis from Fig.~\ref{fig:feature_matrix} is that the dependency between radio channel properties (including IBO) and the resultant SDR can be based on the graphical representation of the feature matrix. The problem is how to map a 2-dimensional feature matrix into the correct SDR value. The ML methods can help in this matter. We propose to treat the feature matrix as an image and utilize image processing techniques for the regression task, which aims to find a mapping between the feature matrix and the SDR. For the image processing tasks, the common approach is to utilize the Convolutional Neural Networks (CNN)~\cite{Nam2016}. While designing an effective CNN architecture from scratch is a non-trivial task, our approach is to utilize a well-established model. One of such CNNs is the so-called VGG16~\cite{simonyan2014very}. It has already been proven to provide high performance in a variety of classification tasks, e.g., vehicle classification~\cite{ma2019}, or spectrum monitoring~\cite{Bhatti2021}. 
The architecture of VGG16 for SDR prediction is depicted in Fig.~\ref{fig:vgg16}. 
Compared to the original VGG16 architecture (see~\cite{simonyan2014very}), the input and output layers are adjusted to the SDR prediction. The input layer is $K\times K$ feature matrix with each element being a real, nonnegative number, while the output is a scalar value of predicted SDR in dB. This is because the original task of VGG16 was classification, while in this paper, we use it for regression following the general idea proposed in~\cite{electronics12183980}. The remaining layers of VGG16 are unchanged. There are 5 \emph{Conv} stages (each utilizing 2 or 3 convolution layers with $3\times 3$ filters, and a max-pool layer), followed by the two fully connected layers of 4096 neurons (\emph{fc6} and \emph{fc7}). For the details, see Fig.~\ref{fig:vgg16}.


\subsection{VGG16 Dastaset and Training}
\begin{figure}[!t]
\centering
\includegraphics[trim={1.5cm 0cm 1.5cm 1.5cm}, clip,width=0.49\textwidth]{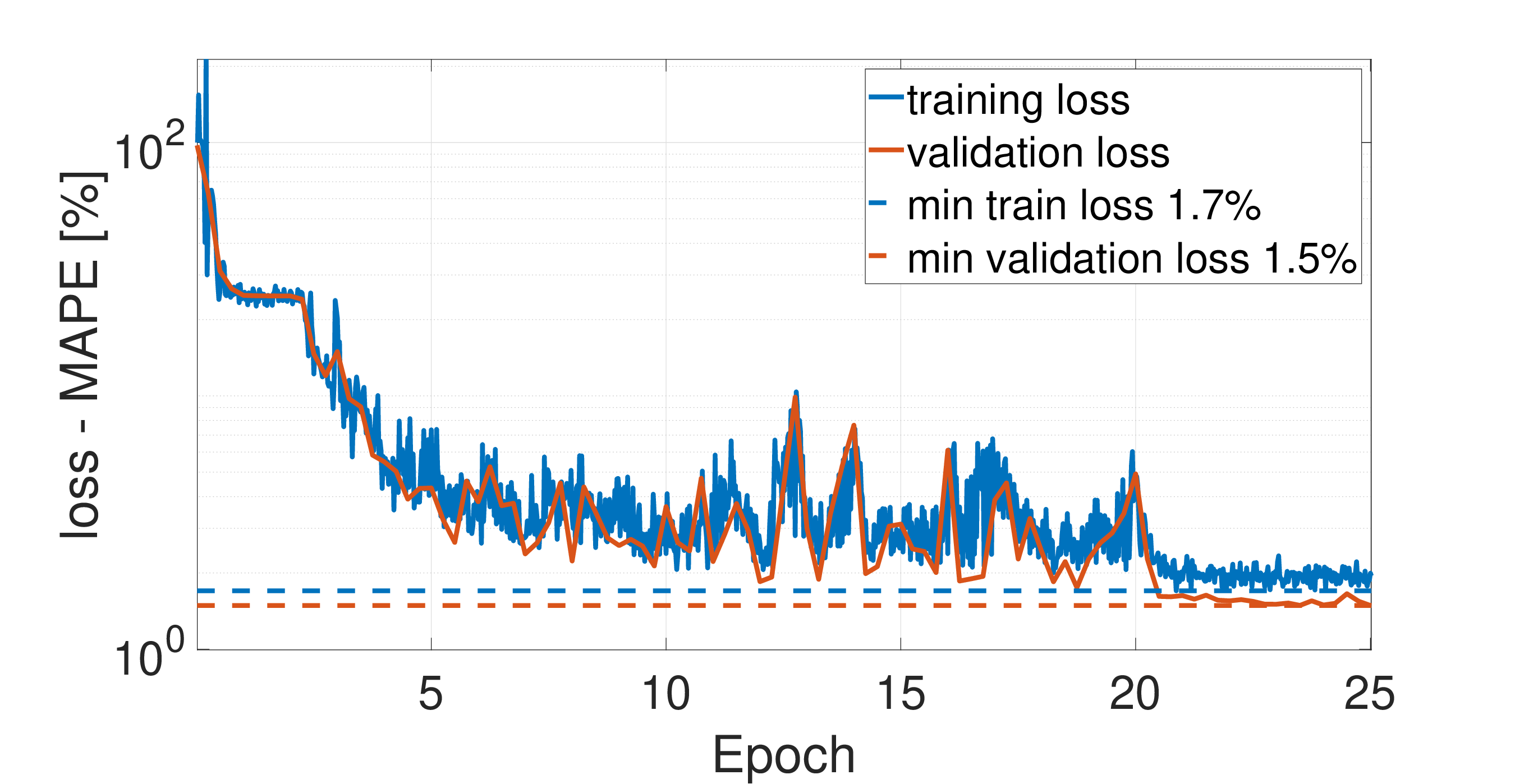}
\caption{Training and validation loss.}
\vspace{0.25cm}
\label{fig:training}
\end{figure}

Using our simulation environment (see Sec.~\ref{subsec:simulation_setup}), we have captured the data to train and evaluate the VGG16 network. The data is obtained using the 3D-RT-based radio channel model. For the training purpose, we used $3542\cdot 4=14168$ feature matrices, i.e., for each of 3542 UE locations, there were feature matrices captured for {\MH IBO $\gamma=\{-3,0,3,6\}$~dB.} Each feature matrix is labeled with a resultant SDR in dB. These data samples are then split into the training and validation sets by randomly selecting $20$\% and $80$\% of samples for validation and training datasets, respectively. {\MH This approach prevents overfitting during the training, i.e., the validation dataset is not used to update VGG16 network, but is used to verify the progress of training on the unknown data.} Next, the test dataset consists of 10626 data samples (feature matrices labeled with SDR), obtained for 3542 UE locations with {\MH IBO $\gamma=\{-1,2,5\}$~dB each}. Although the size of both test and training datasets is smaller compared to, e.g.,~\cite{DEMYTTENAERE201638}, the nature of the feature matrices is more structured than standard images (see examples in Fig.~\ref{fig:feature_matrix}). Therefore, as will be demonstrated through validation and testing, a smaller dataset can provide relatively good performance and generalization. While predicting the SDR, it is important to take into account the relative error, e.g., the absolute error of $3$~dB is more significant when the ground truth is equal to 21~dB than when it's 43~dB. Thus, we decided to utilize the Mean Absolute Percentage Error (MAPE) loss function~\cite{DEMYTTENAERE201638}. We trained the VGG16 network over 25 epochs using the state-of-the-art Adam optimizer with an initial learning rate of $10^{-3}$, and a batch size of 256. The related training and validation loss is depicted in Fig.~\ref{fig:training}. It can be seen that at the beginning, MAPE is significant, i.e., over 100\%. However, it successively decreases to reach the minimum of 1.7\% between the 20th and 25th epoch. The stabilization of the learning curve after the 20th epoch also indicates that the selected training time is sufficient. Most importantly, the validation loss being close to the training loss shows that the model approximates the unknown data well. {\MH To support the reasoning on excluding the information about the phase of the cross-antenna correlation matrix elements presented at the beginning of this section, we repeated the training with an extended feature matrix containing the phase information. In detail, {\MH we used $K\times2K$ feature matrix being a concatenation of original $K\times K$ feature matrix defined in~\eqref{eq:feature_matrix} and $K\times K$ matrix containing the phase of the cross-antenna correlation matrix}. However, after 25 epochs, the MAPE reached about 1.5\%, meaning that no validation or training loss improvement was observed. 
This justifies the definition of the feature matrix of~\eqref{eq:feature_matrix}}.

\begin{figure}[!t]
\centering
\includegraphics[trim={1.5cm 0cm 1.0cm 1.5cm}, clip,width=0.49\textwidth]{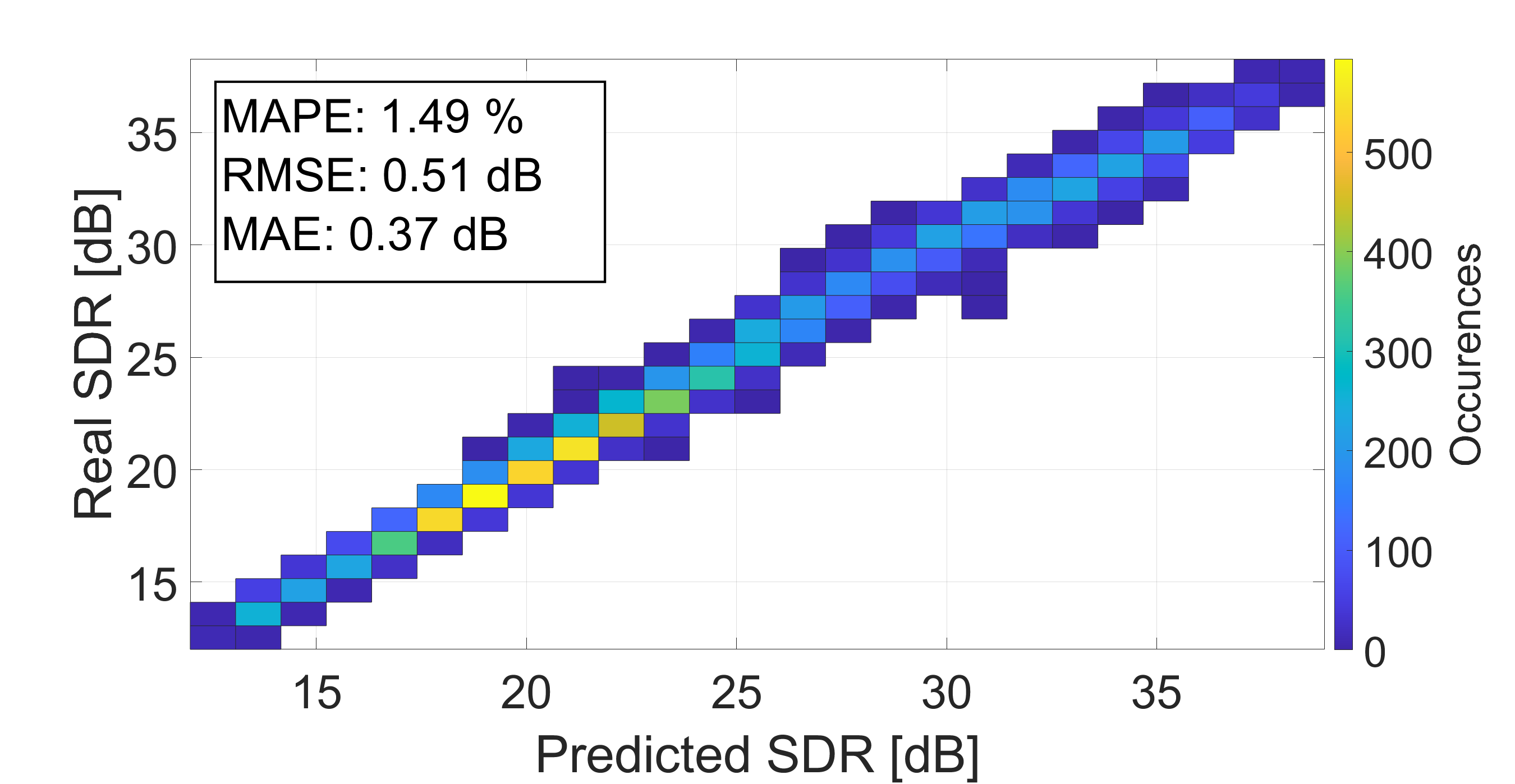}
\caption{Bivariate histogram of predicted and real values of SDR obtained under the test set together with evaluation metrics: MAPE, RMSE, MAE.}
\vspace{0.25cm}
\label{fig:test_results}
\end{figure}

However, to fully evaluate the ability of the VGG16 to predict SDRs on the basis of the feature matrices, we evaluated the trained model against the 
test dataset, exploiting values of IBO different from those used in the training and validation sets, i.e., -1, 2, and 5 dB. The bivariate histogram of real and predicted values of SDR under this test set is presented in Fig.~\ref{fig:test_results}. It can be seen that the relation between predicted and real SDR is almost 1:1, meaning that the trained model can be used to predict SDR under unknown IBO, providing reliable results. Most of the test cases is in the range between about 17 and 25 dB. In addition, in Fig.~\ref{fig:test_results}, there are summarized evaluation metrics for the test set: MAPE, Root Mean Squared Error (RMSE), and Mean Absolute Error (MAE) with the results 1.49\%, 0.51~dB, and 0.37~dB, respectively.

\subsection{Practical Considerations}

The created ML model is specific to a given antenna array geometry and PA properties. As such, we expect that the proposed VGG16 model can be pre-trained and delivered by the vendor of the radio unit or remote radio head. The vendor can obtain the training datasets following one of these strategies:
\begin{itemize}
    \item Utilize the synthetic data approach, i.e., similarly to the approach demonstrated in this paper, the vendor can generate various radio channel characteristics using the 3D-RT. However, instead of the soft-limiter, the vendor can utilize the characteristic of PA specific to a given radio unit, e.g., measured in the lab environment. {\MH To demonstrate the capability of training the proposed VGG16 model against an alternative PA model, we followed the training procedure described in Sec .~\ref {sec:ml_based_sdr_predioction} using the SDR samples collected under the Rapp PA model ($p=2$) defined by~\eqref{eq:rapp_pa}. The performance of the VGG16 against the test dataset is similar to the soft-limiter with MAPE, RMSE, and MAE shown in Tab.~\ref{tab:ml_comparison}. This shows the proposed ML model is able to adapt to the PA characteristics different than the soft-limiter. }
    \item Conduct a field trial with channel measurements connected with alternating the PA's operation point to get live samples of SDR along with the channel matrices. This could also be done on a smaller scale, in a lab environment, utilizing professional radio channel emulators.
\end{itemize}
It is also possible that the specifications of future 6G wireless networks can extend the measurement reporting capabilities of the UEs to report SDR. This could be done using similar mechanisms to these, which are used in 5G to report, e.g., channel state information or SINR. In this case, online learning or tuning of the ML model will be possible
\begin{table}[h]
\caption{Comparison of VGG16 variants}
\label{tab:ml_comparison}
\centering
\begin{tabular}{lccc}
\hline
VGG16 Variant & Soft-limiter & Rapp & Soft-limiter 40\% Pruned \\
\hline
No. parameters & $65 \cdot 10^6$  & $65 \cdot 10^6$ & $39 \cdot 10^6$  \\
No. MACs & $5 \cdot 10^9$ & $5\cdot 10^9$ & $3 \cdot 10^9$ \\
Inference time [ms] &  18.4 & 18.4 & 9.5 \\
MAPE [\%] & 1.49 & 1.25 & 1.51 \\
RMSE [dB] & 0.51 & 0.43 & 0.54 \\
MAE [dB] & 0.37 & 0.31 & 0.38 \\
\hline
\end{tabular}
\end{table}

{\MH Although the results show that the proposed VGG16 provides good SDR estimation, it is characterized by a relatively large number of parameters, e.g., about $65\cdot 10^6$. One common approach to reduce the size of neural networks is to apply pruning, e.g., removing filters (for convolutional layers) and connections (for dense layers) that are associated with weights of the least magnitudes~\cite{He2024}. We pruned the proposed VGG16 model by increasing its sparsity, defined as the ratio between the number of parameters of the initial VGG16 and the number of parameters of the pruned VGG16. After each pruning iteration, the MAPE is calculated using the original validation dataset obtained under the soft-limiter PA model. The results depicted in Fig.~\ref{fig:prunning} show that the size of VGG16 can be reduced by up to 40\% without significant degradation of MAPE, i.e., the number of parameters can be reduced from $65\cdot 10^6$ to $39\cdot 10^6$, while MAPE increases from 1.49 only to 1.51 (see Tab.~\ref{tab:ml_comparison}). The pruning also reduces the number of Multiply and Accumulate operations (MACs) by 40\% as shown in Tab.~\ref{tab:ml_comparison}. This should also decrease the inference time. This has been measured on a 32-core CPU (AMD Ryzen Threadripper PRO 5975WX), and the NVIDIA RTX A5000 GPU. While the original VGG16 has an inference time of $18.4$~ms, after 40\% pruning, it reduces to only $9.5$~ms. 
However, additional hardware optimization, or dedicated acceleration cards, can speed up this process even more. 

}

{\MH 

\begin{figure}[!t]
\centering
\includegraphics[trim={0cm 0cm 0cm 0cm}, clip,width=0.49\textwidth]{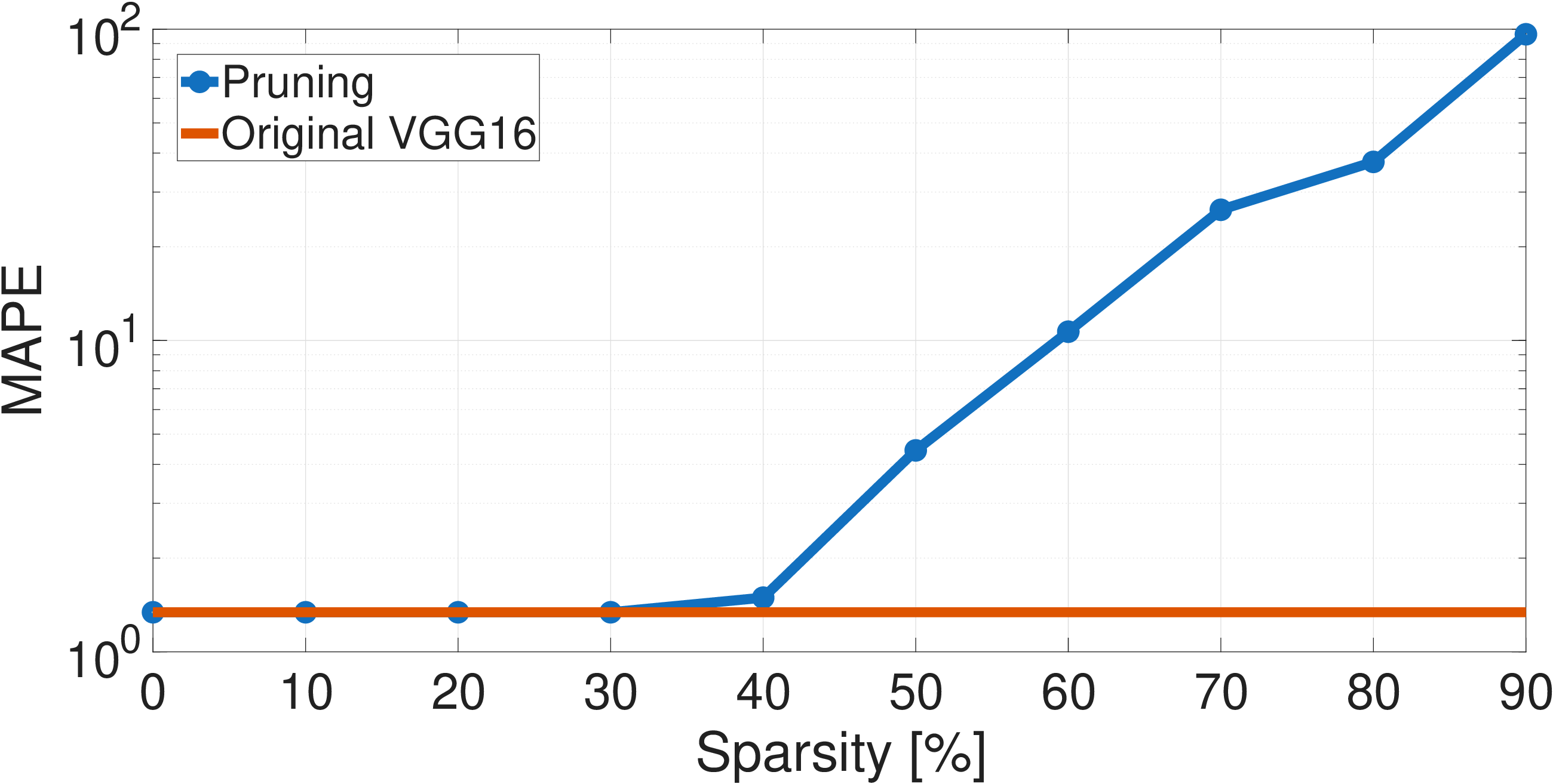}
\caption{ \MH Pruning of the trained VGG16 network.} 
\vspace{0.25cm}
\label{fig:prunning}
\end{figure}
}

\subsection{Distortion-aware M-MIMO per-user power allocation}

We have shown that the VGG16 can be successfully used to map the feature matrix (see \eqref{eq:feature_matrix}) to the SDR. An example use case for this model is the distortion-aware downlink per-user power allocation. This can be thought of as an extension to our previous work, which considered a single scenario of analog beamforming, for which the distortion signal is always directed towards the receiver~\cite{Hoffmann2023}. 
We assume that the radio channel between the $m$-th UE and the MISO BS is known, and the noise plus interference over all subcarriers can be estimated by the UE and is equal to $\sigma^{\mathrm{interf}}$. 
The proposed power allocation can be explained in the following steps, seen from the perspective of the scheduled UE:
\begin{enumerate}
    \item Use radio channel matrix $\mathbf{H}_m$ to create $Z$ feature matrices $\mathbf{F}_m^{(z)}$ by \eqref{eq:feature_matrix} where $z\in \{1,2,..Z\}$. Each $z$ represents different IBO value, equally spaced in the range specific for a given transmitter, i.e., based on the radio channel matrix $\mathbf{H}_m$ reported by the scheduled UE, create $Z$ feature matrices by applying in~\eqref{eq:feature_matrix} $Z$ different values of {\MH average IBO $\gamma$ defined in~\eqref{eq:ibo}.} 
    \item Utilize VGG16 to predict $SDR^{(z)}$ for each feature matrix $\mathbf{F}_m^{(z)}$.
    \item Use \eqref{eq_wanted_RX_power} to estimate the received signal power $S^{\mathrm{RX}, (z)}$ for each {\MH IBO $\gamma$}.
    \item for each {\MH IBO~$\gamma$}, on the basis of the predicted $SDR^{(z)}$, calculate the predicted distortion power $\hat{D}^{\mathrm{RX}, (z)}$:
    \begin{equation}
        \hat{D}^{\mathrm{RX}, (z)}=\frac{S^{\mathrm{RX}, (z)}}{SDR^{(z)}}.
    \end{equation}
    \item select the index $\hat{z}$ of {\MH IBO~$\gamma$} that is associated with the highest estimated SNDR:
    \begin{equation}
        \hat{z} = \arg \max_z \left( \frac{S^{\mathrm{RX}, (z)}}{\hat{D}^{\mathrm{RX}, (z)} + \sigma^{\mathrm{interf}}} \right).
    \end{equation}
\end{enumerate}
This procedure aims to maximize the SNDR for a scheduled UE based on the SDR predictions from the VGG16. In this case, this is equivalent to maximization of the throughput.

\begin{figure}[!t]
\centering
\includegraphics[trim={0cm 0cm 0.0cm 0cm}, clip,width=0.49\textwidth]{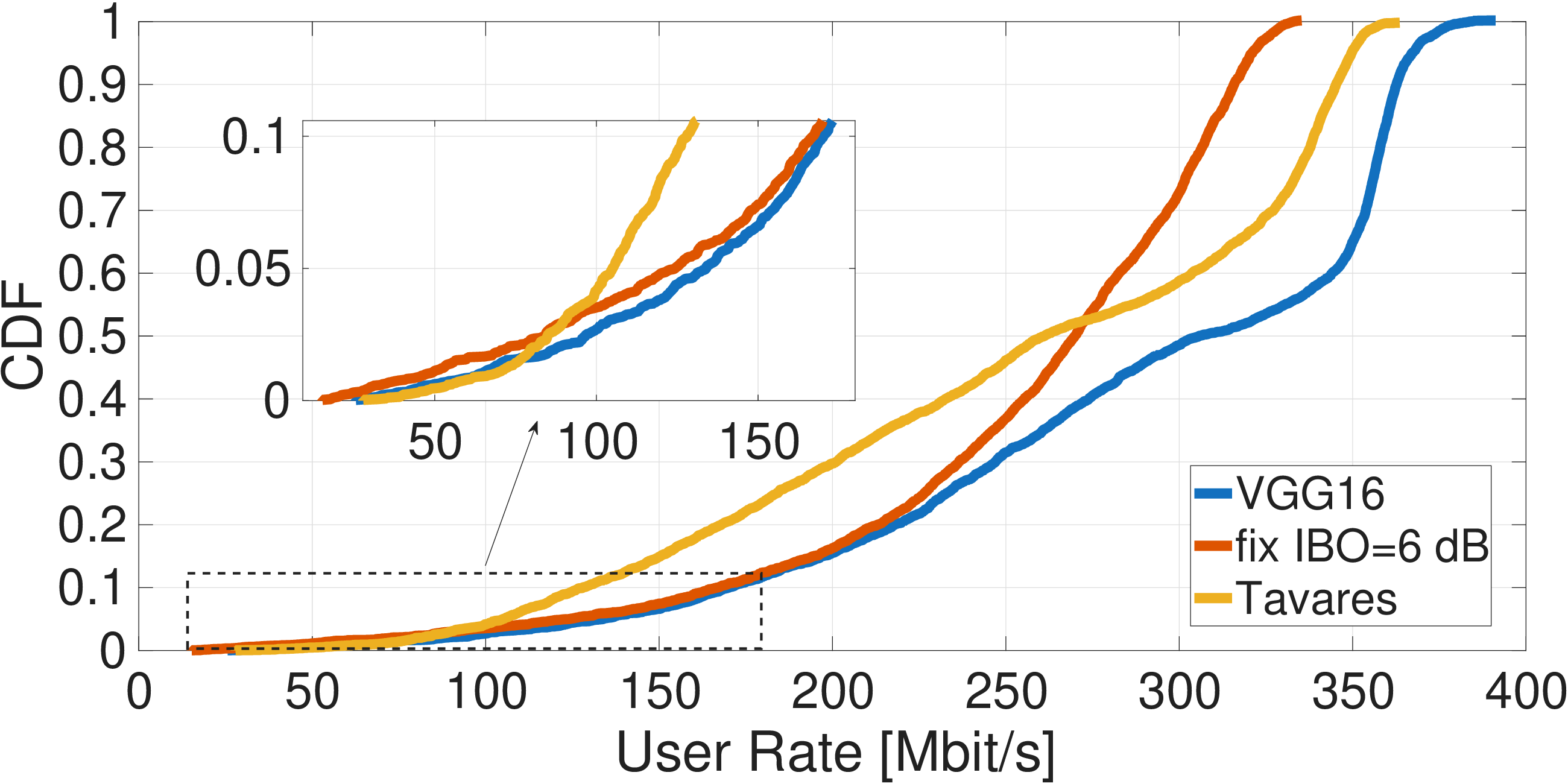}
\caption{\MH CDF of UE rates for the proposed distortion-aware per-user power allocation (\emph{VGG16}), fixed $IBO=6$~dB, and \emph{Tavares}~\cite{tavares2016}.}
\vspace{0.25cm}
\label{fig:rates}
\end{figure}

\begin{figure}[!t]
\centering
\includegraphics[trim={0cm 0cm 0cm 0cm}, clip,width=0.49\textwidth]{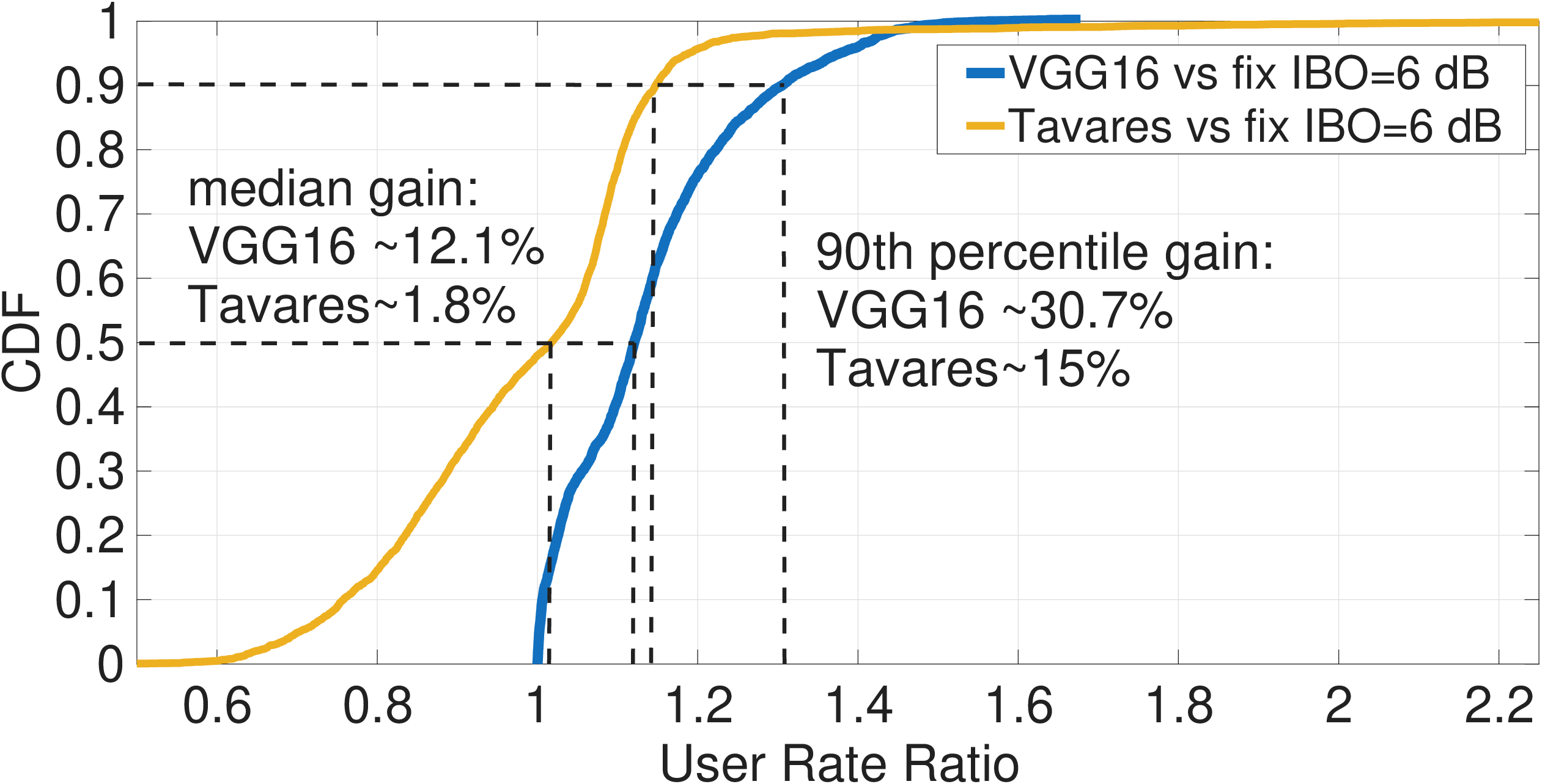}
\caption{ \MH CDF of ratio between the UE rates obtained while using proposed distortion-aware per-user power allocation  or \emph{Tavares}~\cite{tavares2016} and fixed $IBO=6$~dB.}
\vspace{0.25cm}
\label{fig:rates_ratio}
\end{figure}

  To evaluate the proposed distortion-aware power allocation, we utilized the same scenario as described in Sec.~\ref{subsec:simulation_setup}. In addition to the simulation parameters listed therein, we introduced the interference $\sigma^{\mathrm{interf}}=-64$~dBm to mimic the transmission from the adjacent cells. 
  The state-of-the-art approach in mobile networks is to utilize a fixed IBO that ensures low enough nonlinear distortion at the transmitter as discussed in Sec. \ref{sec_introduction}. Following the~\cite{Ahmad2013, Bhat2016} the fixed value of $IBO=6$~dB stands for a good reference value providing sufficiently high wanted signal power while introducing low nonlinear distortion. We compared the fixed $IBO=6~dB$ against the proposed distortion-aware per-user power allocation (denoted on the plots as \emph{VGG16}). The \emph{VGG16} was selecting {\MH IBO~$\gamma$} from the set $\{1,2,\cdots,9\}$~dB. {\MH In addition, we compared the fixed $IBO=6$~dB against the dynamic IBO selection algorithm authored by Tavares et al.~\cite{tavares2016}, which we denote as \emph{Tavares}. The \emph{Tavares} IBO selection algorithm is based on the analytical optimization of IBO for a flat radio channel in a single-antenna BS. We adopted it to the M-MIMO system under the commonly followed assumption of the Rayleigh radio channel.}   
  The CDF of obtained UE rates is depicted in Fig.~\ref{fig:rates}. The first observation is that {\MH compared to the fixed $IBO=6$~dB}, the proposed approach improves the rates, mostly for the users that have a good radio condition, i.e., percentiles above 50, where the gain is in the order of tens of Mbit/s. However, while looking at the zoomed area that corresponds to the percentiles from 0 to 10, we also see that the proposed distortion-aware per-user power allocation outperforms fised IBO solution. {\MH While comparing the proposed \emph{VGG16} approach against the \emph{Tavares}, we see that the achieved UE rates are similar for the percentiles ranging from 0 to about 2. 
  For the higher percentiles the superiority of the proposed solution is clearly visible with the median gain of about 50 Mbit/s.} Next, to capture both gains for UEs characterized by different radio conditions, it is beneficial to assess the ratio between the UE rates obtained while using the proposed distortion-aware per-user power allocation (\emph{VGG16}), {\MH or \emph{Tavares}}, and fixed $IBO=6$~dB. The CDF of such a ratio is shown in Fig.~\ref{fig:rates_ratio}. It can be seen that {\MH for the proposed \emph{VGG16}} the gain ranges from 1 to slightly below 1.7.  It is important to note that no UEs suffer from degraded rates. {\MH On the other hand, for the \emph{Tavares} for a  few UEs the gains can reach 2.2. 
  This is a small fraction of all cases (below 2\%), caused mostly by the prediction error of the \emph{VGG16}.
  Also, in a very few cases, a very low IBO selected by \emph{Tavares}, e.g., $-20$~dB (being out of the set tested by \emph{VGG16}), provides a higher rate than \emph{VGG16}, which for the same UEs selects an IBO of $2$~dB. However, for \emph{Tavares}, about 50\% of UEs suffer rate degradation compared to the fixed IBO scenario, with the worst case degradation of over 40\%.} The results show that, due to the utilization of the proposed distortion-aware per-user power allocation, the median gain in the UE rate is about $12.1$\%, {\MH while the \emph{Tavares} allows for only 1.8\%. Moreover, for the 10\% of UEs (90th percentile), the gain exceeds $30.7$\%, compared to the \emph{Tavares} reaching only about 15\%.}  

\begin{figure}[!t]
\centering
\includegraphics[trim={0cm 0cm 0cm 0cm}, clip,width=0.49\textwidth]{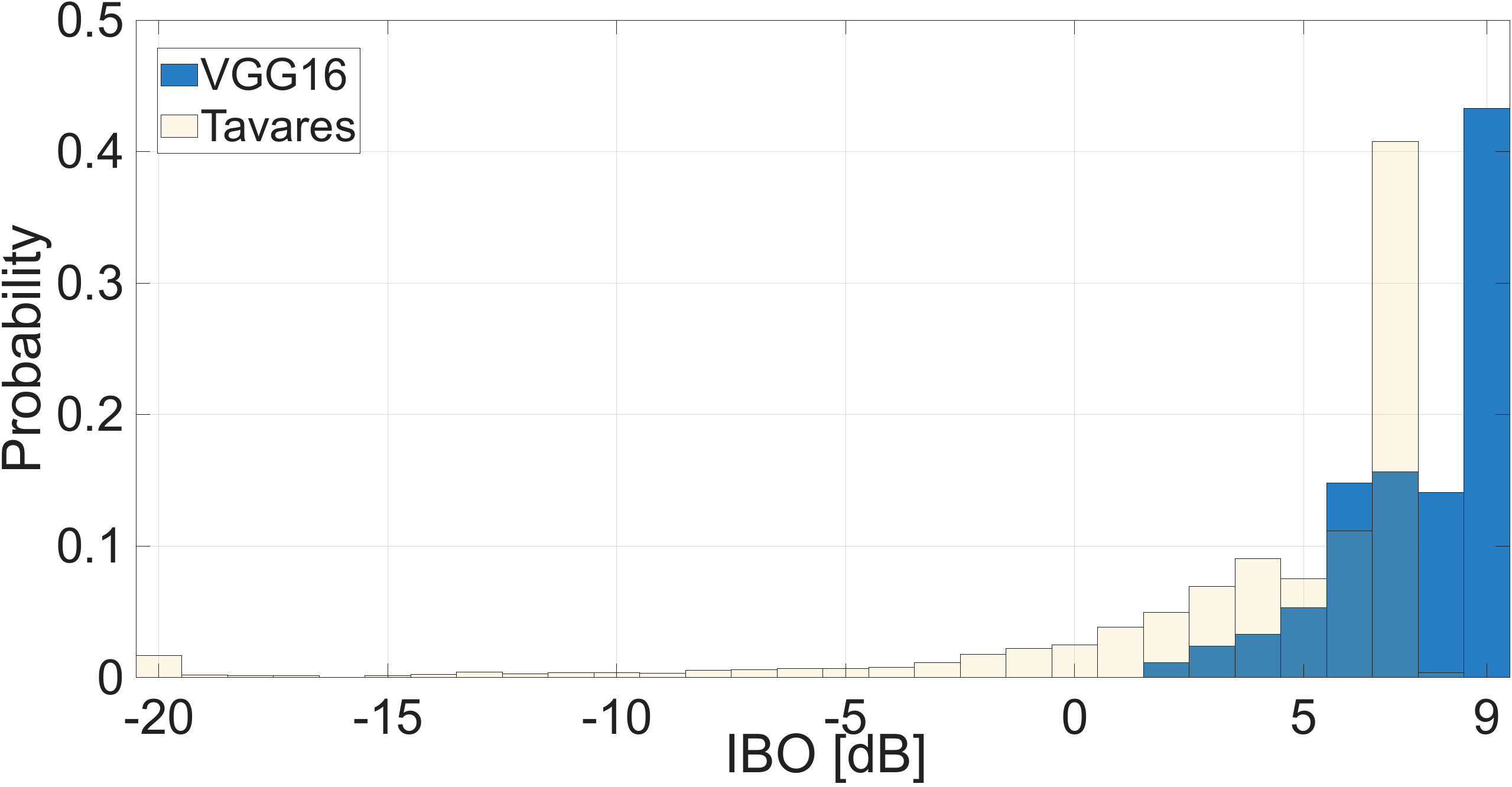}
\caption{\MH Histogram of IBO values selected by the proposed distortion-aware per-user power allocation, and \emph{Tavares}.}
\vspace{0.25cm}
\label{fig:ibo_distribution}
\end{figure}

These results prove the superiority of the proposed distortion-aware per-user power allocation {\MH both over the fixed $IBO=6$~dB, and \emph{Tavares}}. The source of rate gain can be obtained by analyzing the histogram of IBO values selected by the {\MH \emph{Tavares}, and the} proposed distortion-aware per-user power allocation (\emph{VGG16}) as depicted in Fig.~\ref{fig:ibo_distribution}. {\MH For the \emph{VGG16} most often that value of $IBO=9$~dB was selected.} Such a high IBO value is typically selected for UEs of very good radio conditions, for which nonlinear distortion is the main link-quality deterioration factor. In such a case, it is beneficial to reduce the nonlinear distortion power by increasing IBO. As in the considered scenario (see Fig.~\ref{fig:scenario_model_2D}), most of the users are located in the square near the BS, it is reasonable that the probability of selecting $IBO=9$~dB is above 0.4. 
{\MH From this perspective, it can also be seen that the analytical \emph{Tavares} approach based on simplifications on the system model most often selects the lower $IBO=7$~dB. This could be a result of an inaccurate SDR prediction, e.g., omitting the effect of cross-antenna distortion signal correlation at the receiver, leading to the lower UE rates.} On the other hand, for the UEs suffering poor radio conditions it is optimal to increase the wanted signal power at the cost of a rise in nonlinear distortion power. While the noise and inter-cell interference are the main rate-limiting factors, it is optimal to decrease the IBO value by selecting, e.g., 2,3, or 4~dB {\MH indicated by the proposed \emph{VGG16}. However, the \emph{Tavares} due to inaccurate SDR prediction applied much lower values of IBO. In extreme cases, as low as $-20$~dB. In most of the cases, this resulted in a significant nonlinear distortion, which might be the cause of UE rate degradation observed for the \emph{Tavares}.} 

\section{Conclusion} \label{sec:conclusions}
In this paper, we have shown that the nonlinear PAs can have a significant impact on the M-MIMO OFDM system. They generate a nonlinear distortion, acting as an additional source of interference. The literature provided an analysis of these phenomena only for specific radio channels, namely LoS or Rayleigh. Here, the simulation studies based on the realistic 3D-RT radio channel model have shown that neither LoS nor Rayleigh radio channel is a good approximation of the SDR under real-world conditions. When designing the M-MIMO system, the nonlinear distortion should be taken into account both from the perspective of \emph{victim UE} (receiving nonlinear distortion, e.g., from neighboring cells) and \emph{scheduled UE} (receiving nonlinear distortion along with the precoded wanted signal). In the former case, as the radio channels between the BS and UEs from the neighboring cells are usually not known, SDR can be estimated using a statistical model. To estimate SDR, the statistical model uses the theoretical value of SDR for \emph{victim UE}, appropriate for a given IBO. and the GEV distribution, along with the estimated decorrelation distance of approximately 26~m, irrespective of IBO. In the case of \emph{scheduled UE}, the SDR follows a non-trivial distribution and shows a high spatial variability. Moreover, when there is a high variation between the IBO of individual PAs, the observed SDR is below the typically assumed worst case, i.e., LoS scenario. In the case of \emph{scheduled UE}, the radio channel coefficients are known. Thus, it is possible to utilize them to predict SDR. This can be efficiently done using a state-of-the-art CNN ML model - VGG16. To estimate SDR, we proposed to utilize a feature matrix that combines per-antenna IBO and the correlation matrix of the radio channels. The proposed ML model is successfully validated against the test data and used for the PA-aware per-user power allocation. The proposed algorithm optimizes the IBO value in order to maximize the user rate, showing about 12\% median UE gain in user throughput over the fixed IBO scheme of 6~dB. {\MH Moreover, it outperforms the state-of-the-art approach proposed in~\cite{tavares2016}, which is based on the analytical optimization under the simplified flat radio channel and single-antenna BS.} 
 Most importantly, the proposed power allocation algorithm can be treated as an example of how the SDR predicting NN can be used in practice, {\MH e.g., we demonstrate how the proposed methodology can be applied to other PA types, like Rapp model}. As future work, this network can be used, e.g., to improve the energy efficiency of an M-MIMO network or to use some location-awareness and reinforcement learning to find optimal IBO.


%





\ifCLASSOPTIONcaptionsoff
  \newpage
\fi



\bibliographystyle{IEEEtran}
\bibliography{IEEEabrv,bibtex/bib/IEEEexample}
%



%

\begin{IEEEbiography}[{\includegraphics[width=1in,height=1.25in,clip,keepaspectratio]{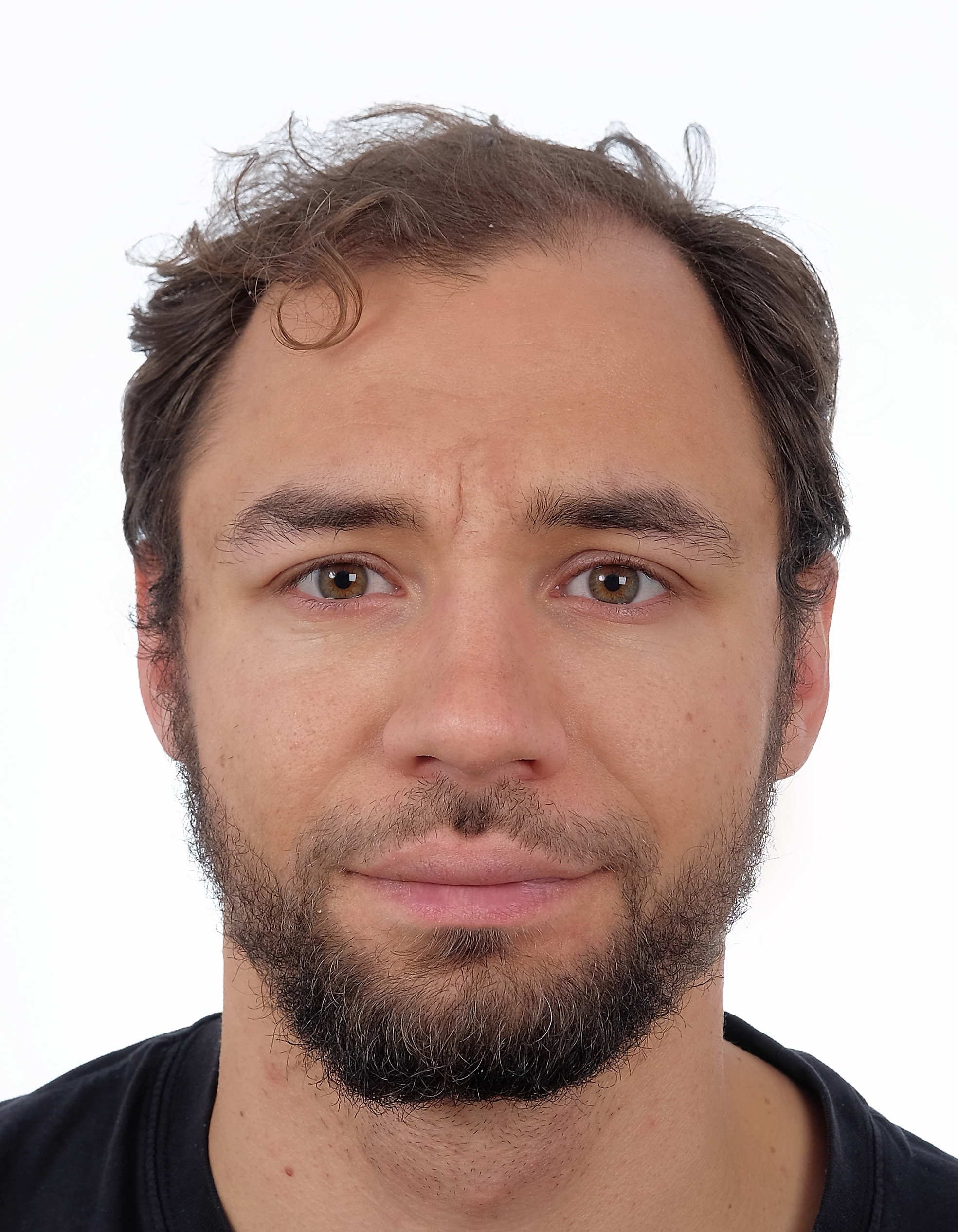}}]{Marcin Hoffmann} (Graduate Student Member, IEEE)  received the M.Sc. degree (Hons.) in electronics and telecommunication from the Poznań University of Technology in 2019, where he is currently pursuing the Ph.D. degree with the Institute of Radiocommunications. He is also a Technical Solution Manager with the Rimedo Labs working on O-RAN software development solutions. His research interests are the utilization of machine learning and location-dependent information for the purpose of network management.
\end{IEEEbiography}
\begin{IEEEbiography}[{\includegraphics[width=1in,height=1.25in,clip,keepaspectratio]{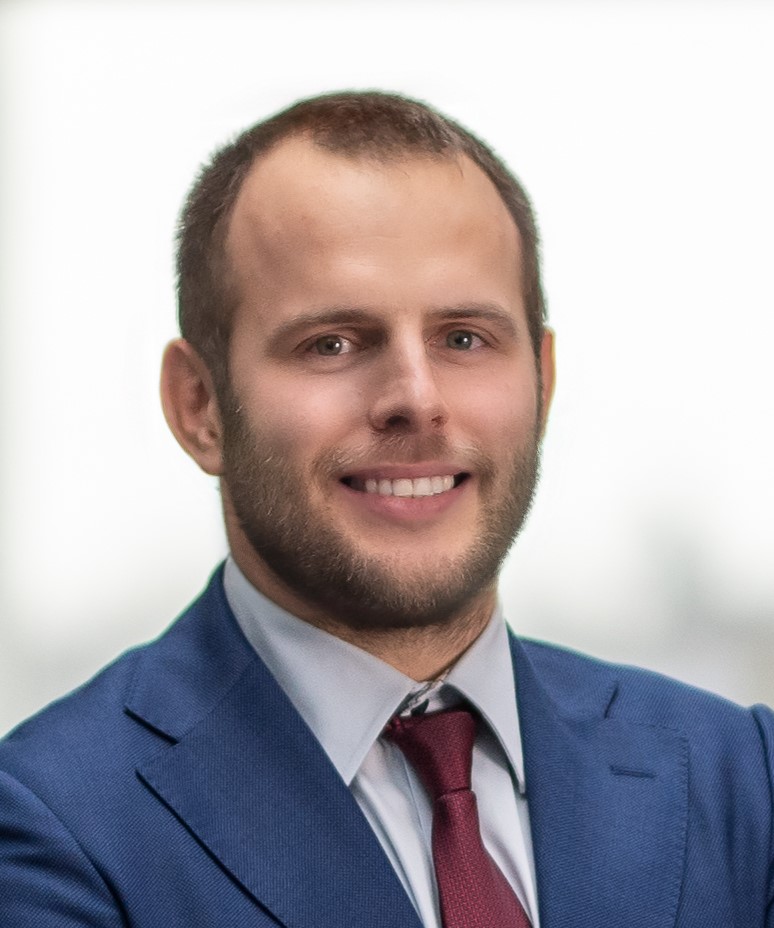}}]{Pawel Kryszkiewicz} (Senior Member IEEE) received the M.Sc. and Ph.D. degrees (Hons.) in telecommunications from the Poznan University of Technology (PUT), Poland, in 2010 and 2015, respectively. He is currently an Associate Professor with the Institute of Radiocommunications, PUT. He was involved in a number of international research projects. His main fields of interest are multicarrier signal design for green communications and problems related to the practical implementation of Massive MIMO systems. 
\end{IEEEbiography}




\end{document}